\documentclass{article}

\usepackage{iclr2027_conference}
\usepackage{iftex}
\ifPDFTeX
    \usepackage{times}
\else
    \usepackage{newtxtext}
\fi
\usepackage{amsmath,amssymb,mathtools,bm,amsthm}
\usepackage{graphicx}
\usepackage{booktabs}
\usepackage{multirow}
\usepackage{float}
\usepackage{flafter}
\usepackage{placeins}
\usepackage{needspace}
\usepackage{wrapfig}
\usepackage{microtype}
\usepackage{xcolor}
\usepackage{tcolorbox}
\usepackage{fontawesome5}
\usepackage[pdfusetitle]{hyperref}
\usepackage{url}
\usepackage{capt-of}

\newcommand{\revision}[1]{#1}

\hypersetup{
    colorlinks=true,
    linkcolor=black,
    citecolor=black,
    urlcolor=blue
}

\definecolor{findingback}{RGB}{243,243,243}
\definecolor{findingframe}{RGB}{72,72,76}
\newtcolorbox{pilotfindingbox}{
    colback=findingback,
    colframe=findingframe,
    boxrule=0.9pt,
    arc=1.5mm,
    boxsep=0pt,
    left=9pt,
    right=9pt,
    top=5pt,
    bottom=5pt,
    before skip=8pt,
    after skip=9pt
}

\newcommand{\pilotfinding}[2]{%
    \par\nopagebreak[4]%
    \begin{pilotfindingbox}
        \textbf{Finding #1:}\enspace\textit{#2}%
    \end{pilotfindingbox}
}

\newcommand{\benchhead}[1]{%
    \makebox[0pt][l]{\rotatebox[origin=bl]{45}{\strut #1}}}

\newtheorem{proposition}{Proposition}

\title{Visual Evidence Prototype-Guided\texorpdfstring{\\}{ }Policy Optimization for Visual Reasoning}
\author{%
    Wenyi Xiao\textsuperscript{1}, Xinchi Xu\textsuperscript{1}, Cong Qian\textsuperscript{1},
    Zexin Hu\textsuperscript{2}, Leilei Gan\textsuperscript{1}, Fei Wu\textsuperscript{1}\\
    \normalfont\textsuperscript{1}Zhejiang University\qquad
    \textsuperscript{2}Ant Group
}
\hypersetup{
    pdfauthor={Wenyi Xiao, Xinchi Xu, Cong Qian, Zexin Hu, Leilei Gan, Fei Wu}
}

\iclrfinalcopy

\begin{document}
\maketitle
\fancyhead[L]{Preprint} 

\begin{abstract}
Reinforcement learning with verifiable rewards has improved visual reasoning in
large vision-language models. Existing methods either rely on visual dependence, while strong visual
dependence does not imply correct use of visual information, or require external
supervision over task-relevant image regions. To address these issues, we introduce \textbf{Visual Evidence Prototypes
(VEPs)}, which derive fine-grained visual supervision from on-policy rollout
outcomes without region annotations. Specifically, for each question, VEPs
aggregate the attention distributions of correct and incorrect rollouts over visual tokens
into Success and Failure prototypes, and contrast them to capture the key
differences. Empirical analyses show that VEPs identify key visual and response tokens for visual reasoning. Intervening on only the top 10\% visual
tokens selected by VEPs reduces answer accuracy by 20.15--28.45 percentage points. Based on these
findings, we propose \textbf{VEP-RL}\footnote{%
    \faGithub\enspace
    \href{https://github.com/Mr-Loevan/VEP-RL}{github.com/Mr-Loevan/VEP-RL}},
which uses VEP to reweight
trajectory-level updates and redistribute token-level
credit to selected tokens. Across fourteen visual reasoning and perception benchmarks, VEP-RL
improves average accuracy over the strongest baseline for each backbone by 2.44,
1.93, and 3.24 absolute percentage points on three models, respectively, with its effectiveness consistently validated in out-of-distribution benchmarks. 
\end{abstract}

\section{Introduction}
\label{sec:introduction}

Reinforcement learning with verifiable rewards (RLVR) has significantly improved the reasoning capabilities of large vision-language models (LVLMs)~\citep{shao2024deepseekmath,deepseekai2025r1,bai2025qwen25vl,wang2025papo,huang2025vppo}.  
By rewarding correct answers, RLVR enables LVLMs to learn from on-policy rollouts without annotated reasoning trajectories~\citep{wang2025vlrethinker,liu2025noisyrollout}. 
However, answer-level rewards alone do not ensure that LVLMs effectively use visual information during reasoning.
Prior studies have shown that models tend to answer questions using textual cues or prior knowledge alone, without access to the image~\citep{chen2024mmstar,zhang2024mathverse}.

To address this limitation, recent work has incorporated additional visual supervision into RLVR~\cite{xiao2025perceptionr1,miao2026prco}.
One line of work estimates visual dependence through image perturbations or internal representations and incorporates these estimates into policy optimization.
For example, PAPO~\citep{wang2025papo} and VPPO~\citep{huang2025vppo} perturb the image and use the resulting changes in token probabilities to measure how strongly the response depends on visual input, while VGPO~\citep{wang2026vgpo} measures the similarity between response and visual representations. 
Such visual dependence guides policy optimization through auxiliary objectives, selective token updates, or advantage reweighting.
However, as \citet{liu2026seeing} observe, strong visual dependence does not indicate that the model uses visual information correctly.

Another growing line of work leverages explicit visual supervision constructed by external tools, human annotators, or stronger models. 
Saliency-R1~\citep{gong2026saliencyr1} aligns model saliency with human-annotated bounding boxes, while Vision-OPD~\citep{yuan2026visionopd} uses an evidence-centered crop as the teacher input. 
OPD-V~\citep{aniri2026opdv} further combines external visual supervision with visual dependence, using a zoomed-in image and a mask image for token-level visual self-distillation. 
Despite their effectiveness, these methods require fine-grained annotations of task-relevant image regions, which we refer to as \textit{visual evidence} in this work.

\begin{wrapfigure}{R}{0.4\textwidth}
    \centering
    \includegraphics[width=\linewidth]{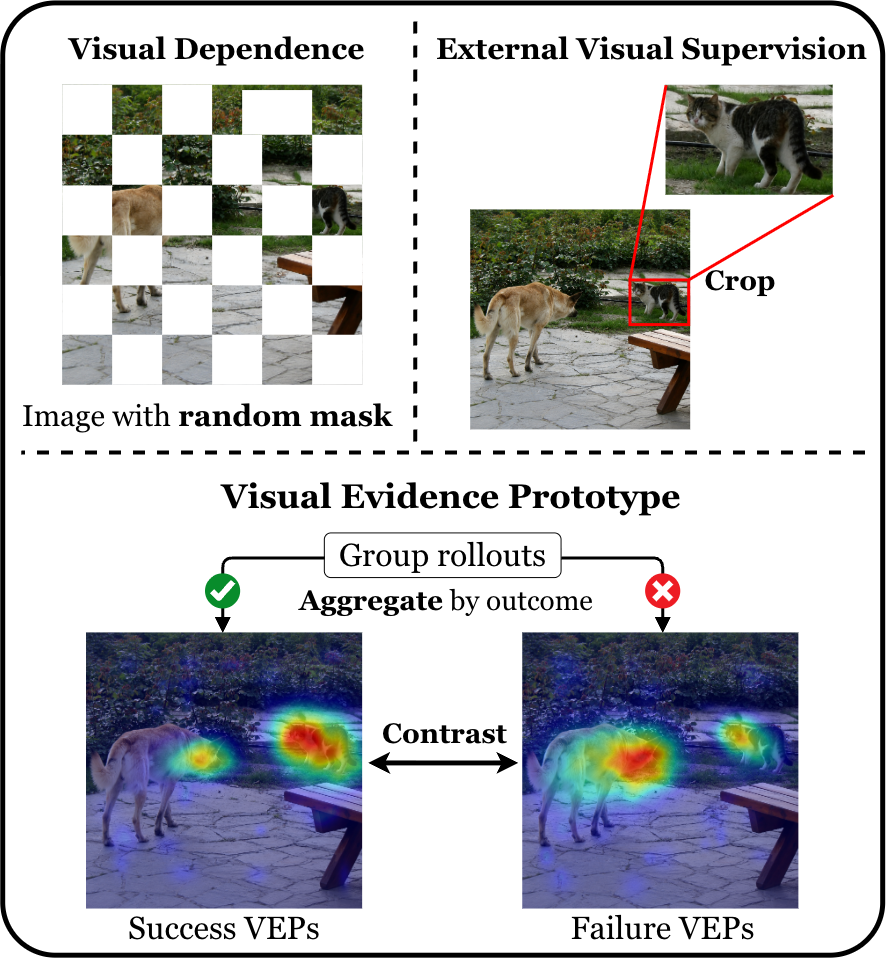}
    \caption{Comparison of visual dependence, external visual supervision, and
    VEPs derived from rollout outcomes.}
    \label{fig:intro_overview}
    \vspace{-1.0\baselineskip}
\end{wrapfigure}

To address these limitations, we propose automatically constructing Visual Evidence Prototypes (VEPs) by aggregating attention patterns from on-policy rollouts according to answer correctness.
For each image--question pair, we partition the rollouts into success and failure groups according to answer correctness.
Within the group, we summarize each rollout's attention distribution over visual tokens and aggregate these distributions to form a Success VEP and a Failure VEP. 
We then contrast the two prototypes to emphasize visual tokens that differ between them. 
As shown in Figure~\ref{fig:intro_overview}, compared with methods that rely on external visual supervision (e.g., Vision-OPD) and visual dependence estimation (e.g., VPPO), VEPs automatically derive supervision from answer-level rewards to localize task-relevant visual evidence, without requiring region or process annotations.

VEPs have demonstrated several characteristics for effective visual supervision across three models:
(i) VEPs can effectively distinguish  independently sampled correct and incorrect rollouts, achieving a mean AUROC of \(0.7127\); 
(ii) Intervening only on the visual tokens (top \(10\%\)) selected by VEPs reduces answer accuracy by \(20.15\)--\(28.45\) percentage points;
(iii) VEPs can be used to localize supporting tokens in correct rollouts and harmful ones in incorrect rollouts.

Based on these characteristics, we propose VEP-RL, which uses VEPs to refine credit assignment in policy optimization at both the trajectory and token levels. 
At the trajectory level, we use each rollout’s relative similarity to the Success and Failure VEPs to determine how strongly to reinforce correct rollouts or penalize incorrect ones. 
At the token level, we intervene on visual tokens selected by the corresponding VEP and use the resulting probability changes to select key response tokens. 
We then redistribute the advantage toward these tokens, strengthening positive updates on supporting tokens and negative updates on harmful tokens. 

Experiments on fourteen visual reasoning and perception benchmarks demonstrate consistent improvements across three models. 
Compared with the strongest baseline for each backbone, VEP-RL improves average accuracy by 2.44, 1.93, and 3.24 absolute percentage points on Qwen2.5-VL-7B, Qwen3-VL-8B, and Qwen3.5-9B, respectively.
On three unseen benchmarks, VEP-RL improves average accuracy over the base models by \(4.29\) and \(1.27\) percentage points on Qwen3-VL-8B and Qwen3.5-9B, respectively. These improvements demonstrate the effectiveness of VEP-RL.


\section{Visual Evidence Prototypes}
\label{sec:vep}

\paragraph{Problem setup.}
For an image \(I\) and a question \(q\), let \(Q=(I,q)\). A vision-language
policy \(\pi\) samples a group of rollouts
\(\mathcal R_Q=\{r_g\}_{g=1}^{G}\), where
\(r_g=(y_{g,t})_{t=1}^{|r_g|}\). Based on answer correctness, the verifier
partitions them into correct rollouts \(\mathcal R_Q^+\) and incorrect
rollouts \(\mathcal R_Q^-\).

Instead of relying on localizing evidence with human annotations, we construct VEPs from correct and incorrect rollouts for the same question. 
We first derive a visual attribution over image tokens for each rollout (\S \ref{sec:rollout_attribution}). 
We then aggregate these attributions by outcome to construct success and failure VEPs, and contrast them to characterize their differences (\S \ref{sec:outcome_prototypes}).

\subsection{Visual Attribution}
\label{sec:rollout_attribution}

To construct VEPs, we first transform each rollout's attention over visual tokens into a \textit{\textbf{visual
attribution}}. Direct attention from the answer to visual
tokens does not account for the attention from intermediate
response tokens, whereas uniformly averaging their visual attention ignores
their different relevance to the answer. We therefore weight each response
token by its connection to the final answer when aggregating its attention
to the visual tokens.

For rollout \(r\), let \(\mathcal I_Q\), \(\mathcal T_r\), and
\(\mathcal A_r\) denote the positions of visual tokens, response tokens
preceding the final answer, and tokens in the generated final answer,
respectively. Let \(A_r(u,v)\) denote the attention to position \(v\) when
predicting token \(u\), averaged over heads and a selected set of decoder
layers \(\mathcal L\). We compute
\begin{equation}
    w_{r,t}
    =
    \frac{\sum_{a\in\mathcal A_r} A_r(a,t)}
    {\sum_{u\in\mathcal T_r}\sum_{a\in\mathcal A_r} A_r(a,u)},
    \qquad
    v_{r,i}
    =
    \frac{\sum_{t\in\mathcal T_r}w_{r,t}A_r(t,i)}
    {\sum_{j\in\mathcal I_Q}\sum_{t\in\mathcal T_r}w_{r,t}A_r(t,j)}.
    \label{eq:rollout_visual_attribution}
\end{equation}
Here, \(w_{r,t}\) measures the relative attention assigned to response
position \(t\) when generating the final answer. We use these weights to
aggregate the response tokens' visual attention and normalize it over
\(\mathcal I_Q\), yielding the rollout-level visual attribution
\(\mathbf v_r=(v_{r,i})_{i\in\mathcal I_Q}\).

\subsection{Visual Evidence Prototypes}
\label{sec:outcome_prototypes}

For the same question, visual attributions can vary across rollouts with the
same outcome. To aggregate the visual patterns for each outcome, we average
the attributions separately over correct and incorrect rollouts:
\begin{equation}
    \boldsymbol{\mu}_Q^+
    =
    \frac{1}{|\mathcal{R}_Q^+|}
    \sum_{r\in\mathcal{R}_Q^+}\mathbf{v}_r,
    \qquad
    \boldsymbol{\mu}_Q^-
    =
    \frac{1}{|\mathcal{R}_Q^-|}
    \sum_{r\in\mathcal{R}_Q^-}\mathbf{v}_r.
    \label{eq:outcome_veps}
\end{equation}
We call \(\boldsymbol\mu_Q^+\) and \(\boldsymbol\mu_Q^-\) the \textbf{\emph{Success
VEP}} and \textbf{\emph{Failure VEP}}, respectively. 
However, visual tokens with high attribution in correct rollouts may also receive high attribution in incorrect ones.
To distinguish the visual evidence emphasized by correct and incorrect rollouts, we construct contrastive VEPs by taking the  differences between the two prototypes:
\begin{equation}
    \mathbf{D}_Q^+
    =
    \boldsymbol{\mu}_Q^+ - \boldsymbol{\mu}_Q^-,
    \qquad
    \mathbf{D}_Q^-
    =
    \boldsymbol{\mu}_Q^- - \boldsymbol{\mu}_Q^+.
    \label{eq:contrastive_veps}
\end{equation}
where subtraction is performed element-wise between attribution values at
corresponding visual-token positions. We call
\(\mathbf{D}_Q^+\) and \(\mathbf{D}_Q^-\) the
\textbf{\emph{Success-Specific Contrastive VEP}} and
\textbf{\emph{Failure-Specific Contrastive VEP}}, respectively. These contrastive VEPs highlight where correct and incorrect rollouts differ in their visual attributions.

\section{Characterizing Visual Evidence Prototypes}
\label{sec:pilots}

To better understand what VEPs capture, we conduct three pilot experiments that examine whether VEPs distinguish rollout outcomes (\S~\ref{sec:pilot_identification}), identify important visual tokens (\S~\ref{sec:pilot_visual_removal}), and localize key response tokens (\S~\ref{sec:pilot_token_intervention}). Detailed pilot settings are provided in Appendix~\ref{app:pilot_details}.

\subsection{VEPs Distinguish Correct and
Incorrect Rollouts }
\label{sec:pilot_identification}

\begin{wrapfigure}{r}{0.50\textwidth}
    \vspace{-0.8\baselineskip}
    \centering
    \includegraphics[width=\linewidth]{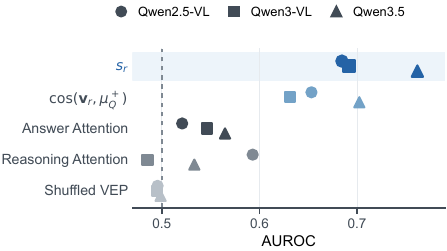}
    \vspace{-0.7\baselineskip}
    \caption{AUROC for distinguishing independently sampled correct and
    incorrect rollouts.}
    \label{fig:pilot1}
\end{wrapfigure}

To investigate whether VEPs provide an informative signal about a rollout's visual reasoning quality, we use answer correctness as a verifiable proxy and test whether VEPs can distinguish correct from incorrect rollouts. Specifically, we construct VEPs from one rollout set and evaluate them on an independently sampled set for each question.
If VEPs capture visual patterns reflecting successful and failed reasoning, correct rollouts should align more closely with the Success VEP relative to the Failure VEP.
We thus define the \textbf{\textit{VEP Similarity Score}} as:
\begin{equation}
    s_r
    =
    \operatorname{cos}\!\left(\mathbf v_r,\boldsymbol\mu_Q^+\right)
    -
    \operatorname{cos}\!\left(\mathbf v_r,\boldsymbol\mu_Q^-\right).
    \label{eq:pilot1_score}
\end{equation}
where \(\operatorname{cos}\) denotes cosine similarity.
Higher scores indicate stronger alignment with visual patterns related to successful reasoning.
As shown in Figure~\ref{fig:pilot1}, VEP Similarity Score achieves a mean AUROC of $0.7127$ across the three models, compared with $0.6624$ using similarity to the Success VEP alone.
Answer Attention, the total attention to visual tokens averaged over answer tokens achieves $0.5441$.
Reasoning Attention, the total attention to visual tokens averaged over response tokens, reaches $0.5373$.
Shuffling the outcome labels used to construct VEPs reduces AUROC to $0.4967$.
Together, these results suggest that VEPs may signal a rollout's visual reasoning quality.

\par\WFclear

\pilotfinding{1}{Success and Failure VEPs capture visual patterns
that distinguish correct and incorrect rollouts across independent samples.}

\subsection{Contrastive VEPs Identify Supporting Visual Tokens}
\label{sec:pilot_visual_removal}

\begin{wrapfigure}{R}{0.50\textwidth}
    \vspace{-0.8\baselineskip}
    \centering
    \includegraphics[width=\linewidth]{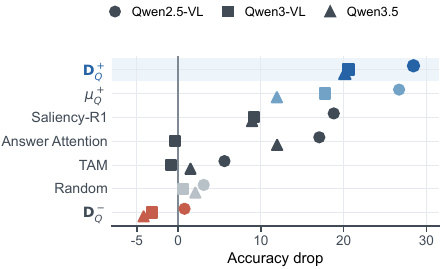}
    \vspace{-0.7\baselineskip}
    \caption{Accuracy drop after replacing \(10\%\) of the visual-token
    hidden states selected by each method. All methods use the same
    intervention setting across three models.}
    \label{fig:pilot2}
    \vspace{-1.0\baselineskip}
\end{wrapfigure}
Pilot \ref{sec:pilot_identification} shows that VEPs capture visual patterns that distinguish correct and
incorrect rollouts. We next examine whether the visual tokens highlighted by
these patterns support correct answers. Specifically, we select \(10\%\) of the visual
tokens under each method and replace their hidden states with the mean visual state of the same
image and resample a complete response. Under the same intervention setting, we
compare \(\mathbf D_Q^+\) with the Success VEP
\(\boldsymbol\mu_Q^+\), Saliency-R1~\citep{gong2026saliencyr1}, answer attention, TAM~\citep{li2025tam}, random visual tokens, and
\(\mathbf D_Q^-\).

As shown in Figure~\ref{fig:pilot2}, intervening on only 10\% visual tokens selected by
\(\mathbf D_Q^+\) causes the largest accuracy drop across all three models,
with decreases of 28.45, 20.61, and 20.15 percentage points.
The drops are consistently larger than those caused by random selection and saliency methods, showing that \(\mathbf D_Q^+\) identifies visual
tokens important for correct answers. The Success VEP
\(\boldsymbol\mu_Q^+\) shows a similar effect, indicating that the base
prototype also captures supporting visual tokens. In contrast, intervening on
\(\mathbf D_Q^-\)-selected tokens might cause slight improvement in accuracy, suggesting that \(\mathbf D_Q^-\) may capture visual tokens that hinder reasoning.

\par\WFclear

\pilotfinding{2}{Contrastive VEPs identify supporting visual tokens, and intervening on them significantly reduces answer accuracy across three models.}

\subsection{Contrastive VEPs Locate Supporting and Harmful Response Tokens}
\label{sec:pilot_token_intervention}

\begin{wrapfigure}{r}{0.50\textwidth}
    \vspace{-0.8\baselineskip}
    \centering
    \includegraphics[width=\linewidth]{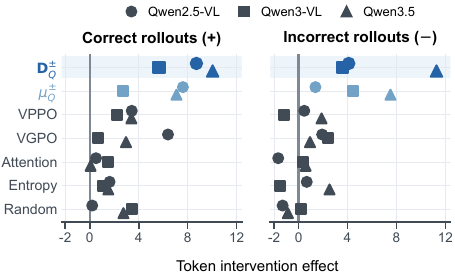}
    \vspace{-0.7\baselineskip}
    \caption{Effect on answer accuracy when the selected response token is
    retained or suppressed. \(\mathbf D_Q^\pm\) and
    \(\boldsymbol\mu_Q^\pm\) denote contrastive and base VEPs.}
    \label{fig:pilot3}
    \vspace{-0.5\baselineskip}
\end{wrapfigure}

Pilot~\ref{sec:pilot_visual_removal} shows that VEPs identify visual tokens
that are important for correct answers. We further ask whether VEPs can also
identify key response tokens in correct and incorrect rollouts. For each
rollout, we intervene on the visual tokens selected by \(\mathbf D_Q^+\) for
correct rollouts and \(\mathbf D_Q^-\) for incorrect rollouts, and measure how
the probability of each observed response token changes under the same prefix. 

Specifically, let \(p_{r,t}\) and \(\widetilde p_{r,t}(\mathbf D)\) denote the probabilities assigned to the observed token at position \(t\) in rollout \(r\) under the original and intervened forward passes, respectively. We define
\begin{equation}
    \Delta_{r,t}(\mathbf D)
    =
    \log p_{r,t}
    -
    \log \widetilde p_{r,t}(\mathbf D).
    \label{eq:token_logp_change}
\end{equation}
A larger positive \(\Delta_{r,t}\) indicates that the selected visual tokens
have a stronger influence on \(y_{r,t}\).
For each rollout, we select the response token with the largest positive
\(\Delta_{r,t}\), excluding final-answer tokens. We then compare two
generations from the same prefix, where the selected token is retained in one and suppressed in the other. For correct rollouts, we compute
\(\operatorname{Acc}(\mathrm{retain})-\operatorname{Acc}(\mathrm{suppress})\), where
a positive value indicates a supporting token. For incorrect rollouts, we
compute the reverse difference,
\(\operatorname{Acc}(\mathrm{suppress})-\operatorname{Acc}(\mathrm{retain})\), where
a positive value indicates a harmful token. We apply the same intervention to
tokens selected by VPPO~\citep{huang2025vppo}, VGPO~\citep{wang2026vgpo},
reasoning attention, entropy, and random selection. 

Figure~\ref{fig:pilot3} shows intervention effects for the two outcomes across all
three models. \textbf{\textit{Suppressing the \(\mathbf D_Q^+\)-selected token decreases answer accuracy
for correct rollouts, whereas suppressing \(\mathbf D_Q^-\)-selected token raises accuracy for incorrect
rollouts}}. Compared with baseline selectors,
contrastive VEPs produce larger accuracy changes in each model and outcome, showing that \(\mathbf D_Q^+\) and \(\mathbf D_Q^-\) locate supporting and
harmful response tokens, respectively.

\par\WFclear

\pilotfinding{3}{Contrastive VEPs locate supporting tokens in correct rollouts
and harmful tokens in incorrect rollouts, even when final-answer tokens are
excluded.}

Together, the three pilots show that VEPs distinguish correct and incorrect rollouts, identify supporting visual tokens, and locate supporting or harmful response tokens. These findings motivate VEP-guided policy optimization at both the trajectory and token levels.

\section{Visual Evidence Prototype-Guided Policy Optimization }
\label{sec:method}

\begin{figure}[!t]
    \centering
    \includegraphics[width=\textwidth]{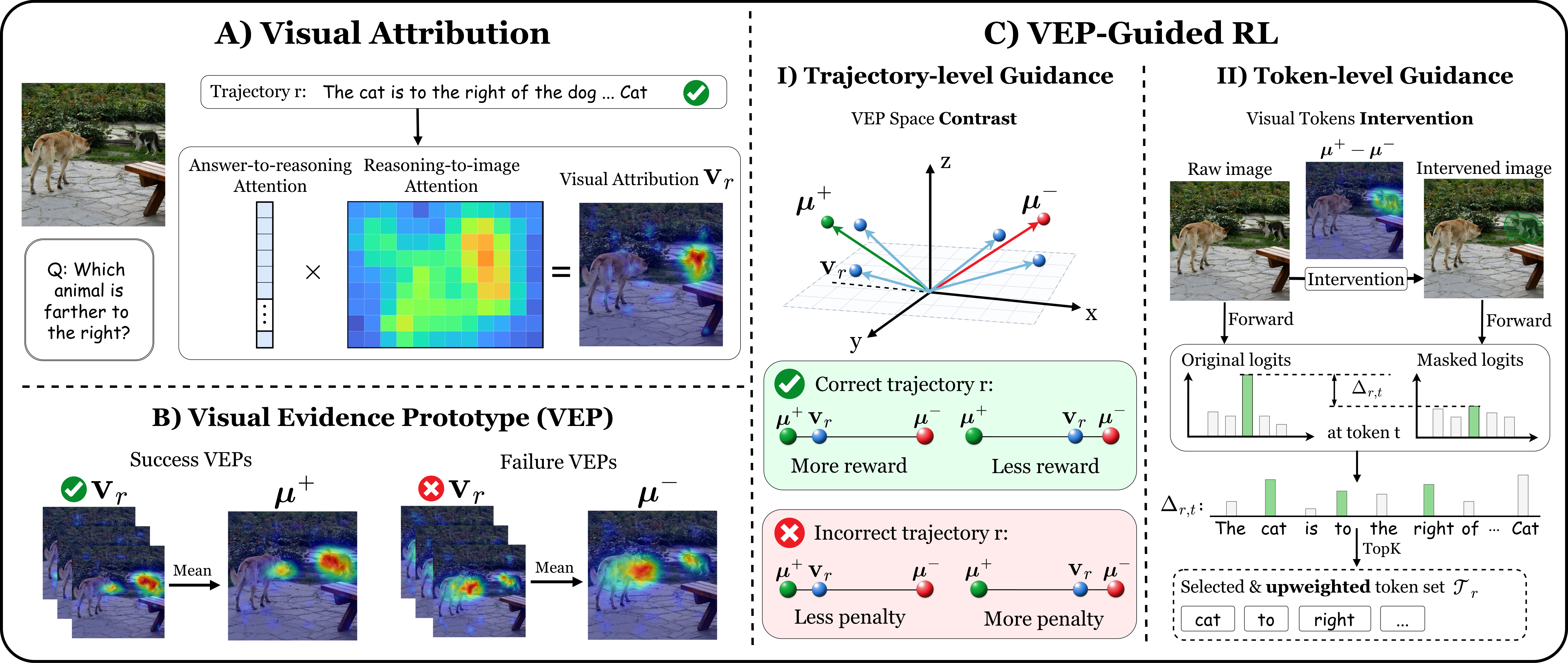}
    \caption{Overview of VEP-RL.
A) Visual attribution for each rollout derived from answer-reasoning-image attention.
B) Correct and incorrect rollouts are aggregated into Success and Failure VEPs.
C) VEP-RL uses VEP Similarity Score for trajectory-level reweighting and contrastive VEPs for token-level credit redistribution.
}
    \label{fig:method_overview}
\end{figure}

Based on the pilot findings, we propose VEP-RL, as illustrated in
Figure~\ref{fig:method_overview}.
VEP-RL uses the VEP Similarity Score to reweight trajectory-level updates and VEP-selected response tokens to redistribute credit within each rollout. We build VEP-RL on Group Relative Policy Optimization (GRPO)~\citep{shao2024deepseekmath}, which estimates
each rollout's advantage relative to other rollouts sampled for the same
input.
The advantage determines the direction and strength of the policy update.
For rollout \(r_i\in\mathcal R_Q\) with
verifiable reward \(R_i\), its normalized advantage is
\(\widehat A_i=(R_i-\operatorname{mean}(\{R_j\}_{j=1}^{G}))/
\operatorname{std}(\{R_j\}_{j=1}^{G})\), which is applied uniformly to tokens within a rollout.


In particular, to construct each outcome prototype from multiple rollouts, VEP-RL applies VEP guidance only
when a rollout group
contains at least two correct and two incorrect rollouts, i.e.,
\(|\mathcal R_Q^+|\geq 2\) and \(|\mathcal R_Q^-|\geq 2\). For all other
groups, we use the original GRPO update objective.

\paragraph{Trajectory-level guidance.}
Finding 1 shows that the VEP Similarity Score distinguishes correct and incorrect rollouts and may signal a rollout’s visual reasoning quality. We therefore use this score to refine GRPO's trajectory-level update strength from a visual supervision perspective. Specifically, we standardize VEP Similarity Score \(s_i\) separately within \(\mathcal R_Q^+\) and
\(\mathcal R_Q^-\) to obtain \(\widehat s_i\), and define:
\begin{equation}
    \widehat A_i^{\mathrm{traj}}
    =
    \begin{cases}
    \left[1+\lambda_{\mathrm{traj}}
    \operatorname{clip}(\widehat s_i,-1,1)\right]\widehat A_i,
    & r_i\in\mathcal R_Q^+,\\[2mm]
    \left[1-\lambda_{\mathrm{traj}}
    \operatorname{clip}(\widehat s_i,-1,1)\right]\widehat A_i,
    & r_i\in\mathcal R_Q^-.
    \end{cases}
    \label{eq:trajectory_vep_guidance}
\end{equation}
Here, \(\lambda_{\mathrm{traj}}\) controls the trajectory-level weighting strength, and we set \(\lambda_{\mathrm{traj}}=0.1\). Since the multiplier changes only the update magnitude, the GRPO advantage determines its direction. Thus, correct rollouts more similar to the Success VEP are updated more strongly, while incorrect rollouts more similar to the Failure VEP are also updated more strongly.

\paragraph{Token-level guidance.}
Finding~3 shows that contrastive VEPs locate supporting response tokens in
correct rollouts and harmful ones in incorrect rollouts. We therefore use
\(\mathbf D_Q^+\) for correct rollouts and \(\mathbf D_Q^-\) for incorrect
rollouts to intervene on the selected visual tokens, and rank response tokens
by the resulting probability change \(\Delta_{r_i,t}\) defined in
Eq.~\eqref{eq:token_logp_change}. We select the highest-scoring response
tokens and redistribute the trajectory-level advantage toward them.
Specifically, let \(\mathcal U_i\) be all response-token positions. We rank these positions by \(\Delta_{r_i,t}\) and select up to the top
\(5\%\) with positive scores as \(\mathcal K_i\subseteq\mathcal U_i\). We then reweight the trajectory-level advantage at each response position:
\begin{equation}
    \widehat A_{i,t}^{\mathrm{VEP}}
    =
    \left[
    1+\lambda_{\mathrm{tok}}
    \left(
    \mathbf{1}[t\in\mathcal K_i]
    -\frac{|\mathcal K_i|}{|\mathcal U_i|}
    \right)
    \right]\widehat A_i^{\mathrm{traj}} .
    \label{eq:token_vep_guidance}
\end{equation}
Here, \(\lambda_{\mathrm{tok}}\) controls the strength of token-level
redistribution, and we set \(\lambda_{\mathrm{tok}}=0.1\). The multiplier
averages to one over \(\mathcal U_i\), selected tokens receive more weight,
and the remaining tokens receive less.



\paragraph{VEP-RL objective.}
Let $y_{i,t}$ denote the $t$-th token of rollout $r_i$, and let
$\rho_{i,t}=
\pi_\theta(y_{i,t}\mid Q,y_{i,<t})/
\pi_{\theta_{\mathrm{old}}}(y_{i,t}\mid Q,y_{i,<t})$
be its importance ratio. 
Building on the GRPO objective given in
Appendix~\ref{app:training_details}, VEP-RL replaces the
trajectory-level advantage with the token-specific advantage
$\hat A_{i,t}^{\mathrm{VEP}}$ defined above. The resulting objective is
\begin{equation}
\mathcal J_{\mathrm{VEP\text{-}RL}}(\theta)
=
\mathbb E_{Q,\{r_i\}}\!\left[
\sum_{i=1}^{G}\!\sum_{t=1}^{|r_i|}\!
\frac{
\min\!\left\{
\rho_{i,t}\widehat A_{i,t}^{\mathrm{VEP}},
\operatorname{clip}(\rho_{i,t},1-\varepsilon_{\mathrm{low}},
1+\varepsilon_{\mathrm{high}})
\widehat A_{i,t}^{\mathrm{VEP}}
\right\}
-\beta D_{\mathrm{KL}}^{i,t}
}{G|r_i|}
\right].
\label{eq:vep_rl_objective}
\end{equation}
\revision{Here, $\varepsilon_{\mathrm{low}}=\varepsilon_{\mathrm{high}}=0.20$
set the clipping ranges, as in our GRPO baseline,} while
$\beta D_{\mathrm{KL}}^{i,t}$ regularizes the policy toward the
reference model.
Thus, VEP-RL changes how the verified trajectory reward is assigned
across rollouts and tokens \revision{on top of the GRPO training recipe
(Appendix~\ref{app:training_details}).}


\section{Experiments}
\label{sec:experiments}
\begin{table}[!hbtp]
    \centering
    \caption{Main results on 14 benchmarks (accuracy, \%). Best and
    second-best results within each base model are shown in \textbf{bold} and
    \underline{underlined}, respectively.}
    \label{tab:main_results}
    \scriptsize
    \setlength{\tabcolsep}{0.8pt}
    \renewcommand{\arraystretch}{1.0}
    \begin{tabular*}{0.98\textwidth}{@{\extracolsep{\fill}}l*{9}{c}@{\hspace{5pt}}*{5}{c}@{\hspace{5pt}}c@{}}
        \toprule
        \multirow{2}{*}{\textbf{Method}}
        & \multicolumn{9}{c}{\textit{Visual Reasoning}}
        & \multicolumn{5}{c}{\textit{Vision-Centric}}
        & \multirow{2}{*}{\textit{Avg.}} \\
        \cmidrule(lr){2-10}\cmidrule(lr){11-15}
        \rule{0pt}{5.8em}
        & \benchhead{MathVerse}
        & \benchhead{MathVista}
        & \benchhead{DynaMath}
        & \benchhead{MathVision}
        & \benchhead{MMK12}
        & \benchhead{ScienceQA}
        & \benchhead{We-Math}
        & \benchhead{Geo3K}
        & \benchhead{LogicVista}
        & \benchhead{MathVerse-V}
        & \benchhead{CLEVR}
        & \benchhead{HalluBench}
        & \benchhead{VLMsAreBlind}
        & \benchhead{V*}
        & \\
        \midrule

\multicolumn{16}{l}{\textbf{\textit{Base model: Qwen2.5-VL-7B-Instruct}}} \\

NoisyRollout
& 57.52 & \underline{70.80} & 64.35 & 20.16 & 47.70 & 86.35 & 63.16 & \textbf{52.91} & 48.77 & 50.38 & 83.50 & 71.82 & 46.56 & 79.06 & 60.22 \\

VL-Rethinker
& 70.09 & 61.50 & 62.36 & \underline{32.03} & 64.90 & 77.48 & 68.85 & 42.26 & 45.86 & 67.05 & 71.00 & 69.61 & 44.49 & 70.68 & 60.58 \\

PAPO
& 67.25 & 70.50 & 63.04 & 27.97 & 72.75 & 89.58 & 67.47 & 41.26 & 48.32 & 63.46 & 89.50 & \underline{72.24} & 48.14 & 79.06 & 64.32 \\

GRPO
& 70.23 & 68.70 & 65.41 & 28.79 & 74.50 & 89.39 & 71.09 & 42.43 & 48.10 & 67.35 & 86.50 & 69.72 & 45.42 & 79.58 & 64.80 \\

DAPO
& 68.39 & 69.20 & 65.41 & 30.51 & 79.90 & 89.11 & 69.60 & 43.26 & 48.55 & 64.37 & 89.00 & 66.98 & 49.63 & 80.10 & 65.29 \\

VPPO
& \underline{71.42} & 65.50 & \underline{66.28} & 30.13 & 81.35 & \underline{91.65} & 71.90 & 44.43 & \underline{49.22} & 68.88 & 84.50 & 70.03 & \underline{49.65} & 82.20 & 66.22 \\

VAPO
& \textbf{72.66} & 69.40 & 65.93 & 30.58 & 74.25 & 89.86 & \textbf{73.39} & 45.09 & \textbf{49.66} & \underline{69.65} & 88.00 & 70.35 & 46.10 & 82.72 & 66.26 \\

VGPO
& 69.91 & 70.40 & 65.44 & 29.17 & \underline{81.80} & 90.90 & 69.89 & 45.42 & 48.32 & 64.76 & \underline{93.00} & 70.45 & 48.40 & \underline{83.25} & \underline{66.51} \\

Ours
& \textbf{72.66} & \textbf{72.30} & \textbf{67.24} & \textbf{32.16} & \textbf{83.20} & \textbf{91.77} & \underline{72.53} & \underline{49.25} & \textbf{49.66} & \textbf{70.03} & \textbf{93.50} & \textbf{74.13} & \textbf{52.03} & \textbf{84.82} & \textbf{68.95} \\

\midrule

\multicolumn{16}{l}{\textbf{\textit{Base model: Qwen3-VL-8B-Instruct}}} \\

OPD-V
& 59.22 & 69.60 & 51.55 & 32.44 & 60.40 & 91.20 & 62.99 & 28.62 & 40.27 & 55.89 & \underline{92.00} & 72.24 & \textbf{67.73} & 85.34 & 62.11 \\

ContextRL
& 72.71 & 71.00 & 74.80 & 48.19 & 67.60 & 95.14 & 78.45 & 64.06 & 56.15 & 70.80 & 88.50 & 72.34 & 52.37 & \underline{87.43} & 71.40 \\

PRCO
& 79.13 & \underline{75.40} & 76.79 & 44.44 & 72.25 & 93.35 & 78.62 & 65.89 & 57.49 & 75.54 & 82.50 & \textbf{76.45} & 61.24 & 82.72 & 72.99 \\

DAPO
& 81.74 & 70.70 & 76.21 & 53.22 & 77.55 & 95.64 & 82.01 & \underline{68.39} & 60.85 & 78.36 & 89.50 & 75.08 & 52.78 & \textbf{89.01} & 75.07 \\

Octopus
& 82.16 & 73.90 & 77.55 & 53.59 & 77.40 & 96.51 & \underline{84.02} & \underline{68.39} & \underline{63.53} & \textbf{80.66} & 89.00 & 75.92 & 57.65 & 85.34 & 76.12 \\

VPPO
& \underline{83.81} & 69.60 & \underline{78.48} & \underline{55.21} & \textbf{80.60} & \underline{97.05} & 83.97 & 67.89 & \underline{63.53} & 80.20 & 90.50 & 75.29 & 58.20 & 85.34 & \underline{76.41} \\

Ours
& \textbf{84.13} & \textbf{75.60} & \textbf{80.01} & \textbf{55.76} & \underline{79.45} & \textbf{97.45} & \textbf{84.77} & \textbf{74.04} & \textbf{64.43} & \underline{80.58} & \textbf{95.50} & \underline{76.03} & \underline{63.60} & 85.34 & \textbf{78.34} \\

\midrule

\multicolumn{16}{l}{\textbf{\textit{Base model: Qwen3.5-9B}}} \\

OPD-V
& 80.00 & 70.60 & 60.50 & 55.62 & 83.60 & 92.45 & 83.33 & 43.09 & 55.03 & 72.78 & 77.50 & 70.24 & 80.80 & \underline{93.19} & 72.77 \\

Vision-OPD
& 83.58 & 71.60 & 63.99 & 58.55 & 85.80 & 92.41 & 88.05 & 46.26 & 67.79 & 75.99 & 87.50 & 71.92 & 85.08 & 91.62 & 76.44 \\

VAD
& 88.94 & 75.30 & 74.41 & \underline{63.67} & 86.85 & 95.05 & \textbf{91.15} & 67.89 & \textbf{77.40} & 86.47 & 86.00 & 76.34 & 84.84 & \textbf{94.76} & 82.08 \\

GRPO
& 89.27 & \underline{76.10} & \underline{86.85} & 61.30 & 86.65 & \underline{98.44} & 87.93 & 84.86 & 70.92 & 85.93 & \underline{95.50} & \textbf{77.92} & \underline{88.49} & 88.48 & 84.19 \\

DAPO
& \underline{89.82} & 75.50 & 86.58 & 63.54 & \underline{92.25} & 97.12 & \underline{88.51} & \underline{87.02} & 72.71 & \underline{87.69} & 95.00 & 75.81 & 87.57 & 85.86 & \underline{84.64} \\

Ours
& \textbf{92.57} & \textbf{76.80} & \textbf{89.91} & \textbf{72.65} & \textbf{94.10} & \textbf{98.89} & \textbf{91.15} & \textbf{90.85} & \underline{75.17} & \textbf{89.98} & \textbf{97.00} & \underline{77.29} & \textbf{92.90} & 91.10 & \textbf{87.88} \\

        \bottomrule
    \end{tabular*}%
    \par\vspace{2pt}
\end{table}

\subsection{Experimental Setup}
\label{sec:experimental_setup}

\paragraph{Training Setup} We train \mbox{Qwen2.5-VL-7B-Instruct}, \mbox{Qwen3-VL-8B-Instruct}, and
\mbox{Qwen3.5-9B} with VEP-RL on ViRL-39K~\citep{wang2025papo}. Our comparisons include
standard RL methods, GRPO and DAPO~\citep{shao2024deepseekmath,yu2025dapo};
visual RLVR methods, including PAPO, VPPO, VGPO, VAPO, NoisyRollout,
VL-Rethinker, Octopus, PRCO, and
ContextRL~\citep{wang2025papo,huang2025vppo,wang2026vgpo,tian2025vapo,
liu2025noisyrollout,wang2025vlrethinker,ding2026octopus,miao2026prco,
lu2026contextrl}, and visual OPD methods, Vision-OPD, VAD, and
OPD-V~\citep{yuan2026visionopd,zhang2026vad,aniri2026opdv}. 
The training settings for our GRPO, DAPO, and VEP-RL runs are detailed in Appendix~\ref{app:training_details}.


\paragraph{Evaluation Setup} We evaluate on
visual reasoning benchmarks including MathVerse, MathVista, DynaMath,
MATH-Vision, MMK12, ScienceQA, We-Math, Geometry3K, and
LogicVista~\citep{zhang2024mathverse,lu2024mathvista,zou2025dynamath,
wang2024mathvision,meng2025mmeureka,lu2022scienceqa,qiao2025wemath,
lu2021intergps,xiao2024logicvista}. We further evaluate vision-centric
reasoning on the vision-dependent subset of MathVerse, CLEVR Counting,
HallusionBench, VLMsAreBlind, and V*Bench~\citep{zhang2024mathverse,
johnson2017clevr,guan2024hallusionbench,rahmanzadehgervi2024blind,
wu2024vstar}. We report accuracy on each benchmark and their macro-average.
Evaluation settings and benchmark details are provided in
Appendix~\ref{app:evaluation_details}.

\subsection{Main Results}
\label{sec:main_results}

\paragraph{Performance across Model Backbones.}
VEP-RL achieves the highest average accuracy on all three model backbones
(Table~\ref{tab:main_results}). Compared with the strongest baseline,
it improves average accuracy by \(2.44\), \(1.93\), and \(3.24\) percentage
points on Qwen2.5-VL, Qwen3-VL, and Qwen3.5, respectively. On Qwen3.5,
VEP-RL further exceeds VAD, the strongest visual OPD baseline, by \(5.80\)
points.

\paragraph{Visual Reasoning and Vision-Centric Performance.}
VEP-RL consistently improves both benchmark groups across all three backbones,
with gains of \(2.10/2.93\), \(1.72/2.30\), and \(3.23/2.39\) points in
visual reasoning and vision-centric accuracy, respectively. The gains span
perception-focused benchmarks such as CLEVR and VLMsAreBlind and visual
mathematical reasoning benchmarks such as Geo3K and MathVision. We report the statistical analysis of Qwen3.5-9B results in Appendix~\ref{app:statistical_analysis}.

\subsection{Ablation Studies}
\label{sec:ablations}

\begingroup
\setlength{\intextsep}{0pt}
\setlength{\abovecaptionskip}{5pt}
\setlength{\belowcaptionskip}{5pt}

\begin{table}[!ht]
    \centering
    \caption{
        Ablation of VEP construction and VEP-RL guidance on Qwen3.5-9B.
        Category-average accuracy (\%).
        VR and VC denote Visual Reasoning and Vision-Centric, respectively.
        In the attribution column, $A$, $R$, and $V$ denote final-answer,
        reasoning (response), and visual tokens.
        We report more ablations of VEP-RL guidance on Qwen2.5-VL-7B in
        Appendix~\ref{app:qwen25_ablation}.
    }
    \label{tab:component_ablation}

    \small
    \setlength{\tabcolsep}{2.5pt}
    \renewcommand{\arraystretch}{0.98}

    \begin{tabular*}{\linewidth}{
        @{\extracolsep{\fill}}
        l
        ccc
        cc
        cc
        @{}
    }
        \toprule
        &
        \multicolumn{3}{c}{VEP Construction}
        &
        \multicolumn{2}{c}{VEP-RL Guidance}
        &
        \multicolumn{2}{c}{Accuracy (\%)} \\
        \cmidrule(lr){2-4}
        \cmidrule(lr){5-6}
        \cmidrule(l){7-8}

        Variant
        & Attribution
        & Aggregation
        & Contrast
        & Trajectory
        & Token
        & VR
        & VC \\
        \midrule

        GRPO
        & -- & -- & --
        & -- & --
        & 82.48 & 87.26 \\

        DAPO
        & -- & -- & --
        & -- & --
        & 83.67 & 86.39 \\

        \addlinespace[3pt]
        \multicolumn{8}{@{}l}{
            \textbf{\textit{VEP construction ablations}}
        } \\[-1pt]

        \multicolumn{8}{@{}l}{
            \hspace{0.8em}\textit{Attribution}
        } \\[-1pt]

        \quad\quad Answer attention
        & $A\!\rightarrow\!V$
        & Outcome mean
        & $\checkmark$
        & $\checkmark$
        & $\checkmark$
        & 85.18 & 87.96 \\

        \quad\quad Reasoning attention
        & $R\!\rightarrow\!V$
        & Outcome mean
        & $\checkmark$
        & $\checkmark$
        & $\checkmark$
        & 85.70 & 88.47 \\

        \addlinespace[2pt]

        \multicolumn{8}{@{}l}{
            \hspace{0.8em}\textit{Aggregation}
        } \\[-1pt]

        \quad\quad Single rollout
        & $A\!\rightarrow\!R\!\rightarrow\!V$
        & Single rollout
        & $\checkmark$
        & $\checkmark$
        & $\checkmark$
        & 85.47 & 88.29 \\

        \quad\quad Leave-one-out VEP
        & $A\!\rightarrow\!R\!\rightarrow\!V$
        & LOO mean
        & $\checkmark$
        & $\checkmark$
        & $\checkmark$
        & 86.31 & 89.26 \\

        \addlinespace[2pt]

        \multicolumn{8}{@{}l}{
            \hspace{0.8em}\textit{Contrast}
        } \\[-1pt]

        \quad\quad w/o prototype contrast
        & $A\!\rightarrow\!R\!\rightarrow\!V$
        & Outcome mean
        & --
        & $\checkmark$
        & $\checkmark$
        & 86.14 & 88.98 \\

        \addlinespace[3pt]
        \multicolumn{8}{@{}l}{
            \textbf{\textit{VEP-RL guidance ablations}}
        } \\[-1pt]

        \quad Trajectory only
        & $A\!\rightarrow\!R\!\rightarrow\!V$
        & Outcome mean
        & $\checkmark$
        & $\checkmark$
        & --
        & 85.52 & 88.63 \\

        \quad Token only
        & $A\!\rightarrow\!R\!\rightarrow\!V$
        & Outcome mean
        & $\checkmark$
        & --
        & $\checkmark$
        & 85.18 & 88.14 \\

        \addlinespace[3pt]
        \textbf{VEP-RL}
        & $A\!\rightarrow\!R\!\rightarrow\!V$
        & Outcome mean
        & $\checkmark$
        & $\checkmark$
        & $\checkmark$
        & \textbf{86.90}
        & \textbf{89.65} \\

        \bottomrule
    \end{tabular*}
\end{table}
\vspace{5pt}
\endgroup

We conduct ablations on both VEP construction and VEP-RL guidance, as shown
in Table~\ref{tab:component_ablation}.
\textbf{First, each component of VEP construction contributes to performance.}
Replacing the full answer--response--visual attribution with answer attention or reasoning attention reduces performance,
while using a single rollout instead of outcome-level aggregation and removing the contrast during token selection also leads to consistent drops.
To examine whether the improvement depends on a rollout contributing to its own
outcome prototype, we further construct a leave-one-out (LOO) variant that
excludes each rollout from the prototype of its corresponding outcome when
computing its VEP guidance.
LOO VEP achieves 86.31 on VR and 89.26 on VC, close to the 86.90 and 89.65 of
the default VEP-RL, showing that the gain is largely preserved without
self-inclusion.
\textbf{Second, both forms of VEP-RL guidance are effective.}
Trajectory-only and token-only guidance both improve over DAPO, while combining
them achieves the best performance, indicating that trajectory- and token-level
guidance provide complementary benefits.
\subsection{Analyses}
\label{sec:analysis}

%

\begin{wrapfigure}{r}{0.6\textwidth}
    \centering
    \includegraphics[width=\linewidth]{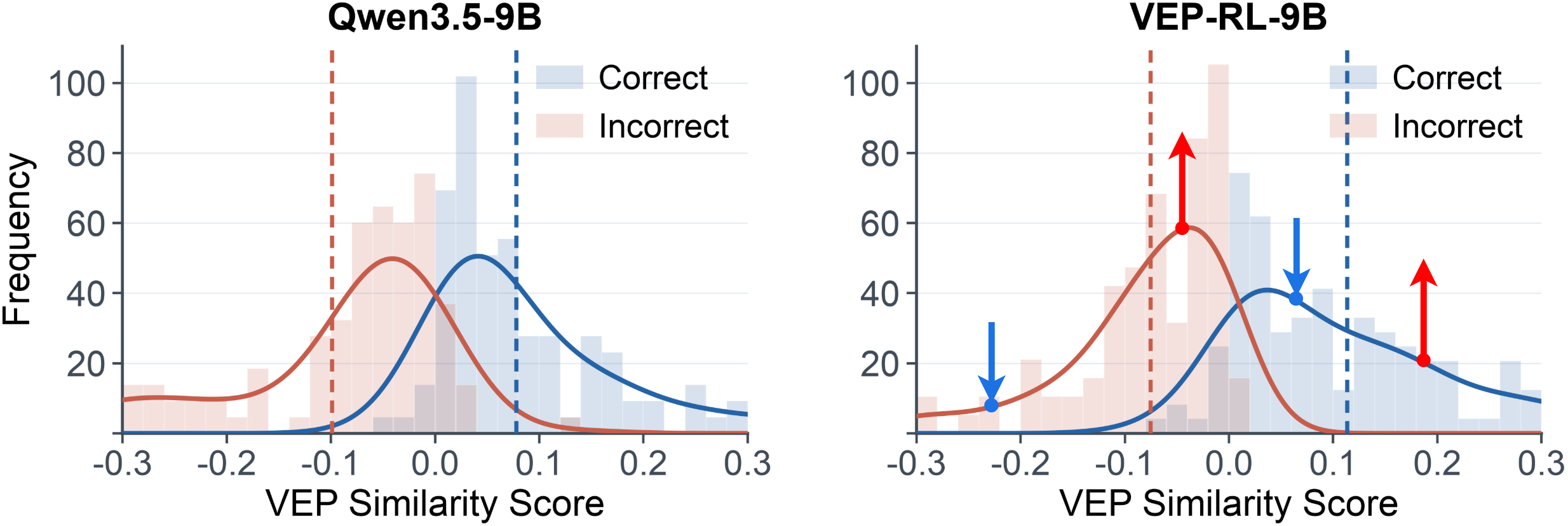}
    \caption{Distribution of the VEP Similarity Score \(s_r\).}
    \label{fig:contrastive_score_before and after training}
\end{wrapfigure}

\paragraph{Analysis of VEP Guidance. }
To investigate the effect of VEP guidance, we measure the VEP Similarity Score of both correct and incorrect rollouts before and after VEP-RL training.
As shown in Figure~\ref {fig:contrastive_score_before and after training}, both distributions shift toward higher scores.
Probability mass in lower similarity score regions is reduced and squeezed into higher similarity score regions within each outcome category.
We call this empirical redistribution the \textbf{VEP Squeeze Effect}. 
This pattern is consistent with our trajectory-level guidance that correct rollouts with higher \(s_r\) receive stronger reinforcement, whereas incorrect rollouts with lower \(s_r\) receive stronger penalties.
This effect, together with the following error analysis in Appendix ~\ref{app:error_analysis}, suggests that our method likely optimizes the model's visual perception capability.



\paragraph{Analysis of Intervention Ratio.}
 \begin{wrapfigure}{t}{0.6\textwidth}
    \vspace{-5pt}
    \centering
    \includegraphics[width=\linewidth]{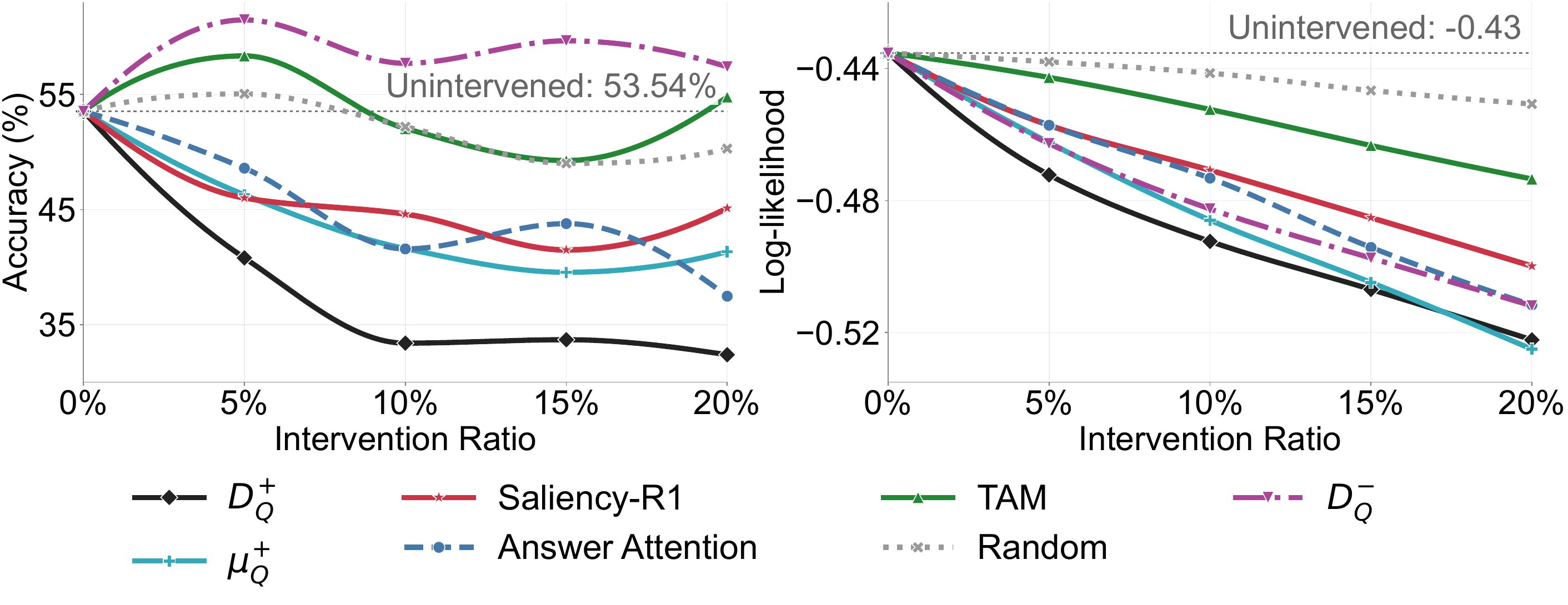}
    \caption{ Answer accuracy and log-likelihood.}
    \label{fig:visual_state_intervention}
    \vspace{-5pt}
\end{wrapfigure}

To examine the choice of the default 10\% intervention ratio, we measure accuracy drops under intervention ratios from 5\% to 20\%.
Beyond accuracy, we also measure the mean token log-likelihood of the original correct response, which reflects the model’s confidence in that response. As shown in Figure~\ref{fig:visual_state_intervention}, intervening on only 5--10\% of the selected visual tokens with \(D_Q^+\) yields significant accuracy drops and a lower likelihood. We therefore adopt 10\% as the intervention ratio in VEP-RL. The effect of VEP intervention becomes marginal beyond 10\%, echoing observations~\citep{yangCVPR2025visionzip} that visual tokens are highly redundant and  \textbf{\textit{useful visual evidence is  sparse}}.
In contrast, random masking is not evidence-aware and removes visual tokens uniformly, often requiring ratios as high as \(60 \%\) or \(80 \%\)~\citep{wang2025papo} to sufficiently intervene on critical visual evidence.






\paragraph{Generalization to Out-of-Domain Tasks.}

\begin{wraptable}{r}{0.50\textwidth}
    \centering
    \caption{Performance on out-of-domain visual reasoning benchmarks.}
    \label{tab:generalization}
    \vspace{0.3\baselineskip}
    \setlength{\tabcolsep}{2.0pt}
    \resizebox{\linewidth}{!}{
    \begin{tabular}{lcccc}
        \toprule
        Model & MMMU-Pro & MM-Star & CV-Bench & Avg. \\
        \midrule
        Qwen3-VL-8B
        & \underline{55.90} & 70.90 & 80.78 & 69.19 \\
        GRPO & 51.97 & 72.13 & 84.95 & 69.68 \\
        VPPO
        & 53.52 & \underline{73.27} & \underline{86.13} & \underline{70.97} \\
        \textbf{VEP-RL-8B}
        & \textbf{58.11} & \textbf{74.87} & \textbf{87.45} & \textbf{73.48} \\
        \midrule
        Qwen3.5-9B
        & \underline{67.40} & \underline{77.13} & \underline{87.87} & \underline{77.47} \\
        GRPO & 62.89 & 76.60 & 87.76 & 75.75\\
        Vision-OPD
        & 57.12 & 72.73 & 85.29 & 71.71 \\
        \textbf{VEP-RL-9B}
        & \textbf{67.51} & \textbf{79.67} & \textbf{89.04} & \textbf{78.74} \\
        \bottomrule
    \end{tabular}}
    \vspace{0.5\baselineskip}
\end{wraptable}

To evaluate whether VEP-RL generalizes beyond the training domains, we
further evaluate the trained models on MMMU-Pro~\citep{yue2025mmmupro}, MM-Star~\citep{chen2024mmstar}, and CV-Bench~\citep{tong2024cambrian}.
As shown in Table~\ref{tab:generalization}, VEP-RL consistently improves both evaluated backbones over their corresponding base models. On Qwen3-VL-8B, VEP-RL
improves the average accuracy from \(69.19\%\) to \(73.48\%\), and also
outperforms VPPO on all three benchmarks. On Qwen3.5-9B, VEP-RL reaches
\(78.74\%\) average accuracy, improving the base model by \(1.27\) percentage
points. It also exceeds Vision-OPD-9B by \(10.39\), \(6.94\), and \(3.75\)
points, respectively. These results show that the improvements of VEP-RL extend to
unseen visual reasoning tasks. More error analysis and case studies of VEP and VEP-RL are provided in Appendix ~\ref{app:additional_analysis}.

\FloatBarrier
\section{Related Work}
\label{sec:related_work}

\paragraph{RLVR for Visual Reasoning.}
Existing RLVR methods improve visual reasoning through better rollout construction, reward design, and optimization objective.
Some methods enrich training rollouts through data augmentation, replay, recombination, or resampling~\citep{liu2025noisyrollout,
wang2025vlrethinker,ding2026octopus,lu2026contextrl,miao2026prco}. 
Some methods complement answer correctness with perception rewards~\citep{xiao2025perceptionr1,tian2025vapo}, saliency-alignment rewards~\citep{gong2026saliencyr1}, or contrastive visual dependency rewards~\citep{huang2026evidencerl}.
FAST-GRPO adapts reasoning depth to question difficulty through length
rewards and difficulty-aware KL regularization~\citep{xiao2025fastgrpo}.
Closer to our work, some methods incorporate perceptual signals into the optimization objective~\citep{wang2025papo,yu2026cfpo}, or leverage visual dependency and visual focus to provide fine-grained optimization signals~\citep{huang2025vppo,wang2026vgpo}, operating at trajectory or token levels.


\paragraph{Visual Supervision Signals.}
Existing methods construct visual supervision from different sources.
Visual dependency measures sensitivity to random perturbations~\citep{wang2025papo,huang2025vppo}, or similarity between response and visual representations ~\citep{wang2026vgpo},
but strong dependence does not necessarily imply correct visual reasoning~\citep{liu2026seeing}.
VL-Calibration combines sensitivity to image perturbations with token entropy
to estimate visual certainty and reweight token advantages~\citep{xiao-etal-2026-vl}.
Explicit guidance from annotated regions depends on the availability and
quality of external evidence~\citep{yuan2026visionopd,liu2026vaopd,sun2026vzero,zhang2026vad,aniri2026opdv}.
For text-to-image alignment evaluation, REVEALER uses a bounding-box
localization reward to train models to ground prompt elements before judging
their alignment with the image~\citep{shi2026revealer}.
Saliency-based signals localize relevant visual regions through attention or activation maps~\citep{li2025tam,gong2026saliencyr1},
but saliency alone does not explicitly distinguish between the visual evidence contributing to successful or failed reasoning processes. 

\paragraph{Preference-Based Alignment.}
Direct preference optimization learns from pairs of preferred and
dispreferred responses~\citep{xiao2026dposurvey}. HSA-DPO uses sentence-level
hallucination feedback to construct such pairs and adapts the preference
objective to hallucination severity~\citep{xiao2025hsadpo}.

\FloatBarrier
\section{Conclusion}
\label{sec:conclusion}

We introduced Visual Evidence Prototypes (VEPs), which derive fine-grained visual supervision from on-policy rollout outcomes without region annotations. 
VEPs aggregate visual attributions from correct and incorrect rollouts into Success and Failure prototypes and contrast them to capture differences in the visual evidence used across outcomes. 
Empirical analyses show that VEPs identify supporting visual and response tokens, and that intervening on only the top 10\% selected visual tokens reduces answer accuracy by 20.15--28.45 percentage points. 
Building on these findings, VEP-RL uses VEP to reweight trajectory-level updates and contrastive VEPs to redistribute token-level credit. 
Across fourteen visual reasoning and perception benchmarks, VEP-RL consistently improves three model backbones over their strongest baselines, with gains also extending to unseen benchmarks.



\subsection*{AI use statement}
Generative AI tools were used to assist with manuscript editing, code
development and debugging, and proving mathematical claims. All
AI-assisted text, code, analyses, and artifacts were reviewed and verified by
the authors. The authors take full responsibility for the final content of
this work.

\subsection*{Ethics Statement}
Our research is conducted entirely within the scope of LVLM reasoning and relies exclusively on publicly available academic benchmarks. These datasets contain no personal, sensitive, or private user information. Our primary objective is to improve AI models' visual reasoning capabilities. Given the focus on abstract problem-solving and the use of established public benchmarks, we are not aware of any direct or foreseeable ethical concerns or negative societal impacts arising from our methodology or findings.

\subsection*{Reproducibility Statement}
To facilitate reproducibility of our work, we describe the experimental setup, including training data, optimization, and rollout configuration, in Appendix~\ref{app:training_details}, and the main training hyperparameters are summarized in Table~\ref{tab:training_hyperparameters}.
Appendix~\ref{app:evaluation_details} describes the evaluation benchmarks, inference settings, metrics, and scoring procedures.
For public baselines, we follow the inference settings in their official repositories. Appendix~\ref{app:pilot_details} describes the pilot-study settings. 

\bibliographystyle{iclr2027_conference}
\bibliography{references}

@article{shao2024deepseekmath,
  title         = {{DeepSeekMath}: Pushing the Limits of Mathematical Reasoning in Open Language Models},
  author        = {Shao, Zhihong and Wang, Peiyi and Zhu, Qihao and Xu, Runxin and Song, Junxiao and Bi, Xiao and Zhang, Haowei and Zhang, Mingchuan and Li, Y. K. and Wu, Y. and Guo, Daya},
  journal       = {arXiv preprint arXiv:2402.03300},
  year          = {2024},
  eprint        = {2402.03300},
  archiveprefix = {arXiv},
}

@article{deepseekai2025r1,
  author = {Guo, Daya and Yang, Dejian and Zhang, Haowei
    and Song, Junxiao and Wang, Peiyi and Zhu, Qihao
    and Xu, Runxin and Zhang, Ruoyu and Ma, Shirong
    and Bi, Xiao and Zhang, Xiaokang and Yu, Xingkai
    and Wu, Yu and Wu, Z. F. and Gou, Zhibin
    and Shao, Zhihong and Li, Zhuoshu and Gao, Ziyi
    and Liu, Aixin and Xue, Bing and Wang, Bingxuan
    and Wu, Bochao and Feng, Bei and Lu, Chengda
    and Zhao, Chenggang and Deng, Chengqi and Ruan, Chong
    and Dai, Damai and Chen, Deli and Ji, Dongjie
    and Li, Erhang and Lin, Fangyun and Dai, Fucong
    and Luo, Fuli and Hao, Guangbo and Chen, Guanting
    and Li, Guowei and Zhang, H. and Xu, Hanwei
    and Ding, Honghui and Gao, Huazuo and Qu, Hui
    and Li, Hui and Guo, Jianzhong and Li, Jiashi
    and Chen, Jingchang and Yuan, Jingyang and Tu, Jinhao
    and Qiu, Junjie and Li, Junlong and Cai, J. L.
    and Ni, Jiaqi and Liang, Jian and Chen, Jin
    and Dong, Kai and Hu, Kai and You, Kaichao
    and Gao, Kaige and Guan, Kang and Huang, Kexin
    and Yu, Kuai and Wang, Lean and Zhang, Lecong
    and Zhao, Liang and Wang, Litong and Zhang, Liyue
    and Xu, Lei and Xia, Leyi and Zhang, Mingchuan
    and Zhang, Minghua and Tang, Minghui and Zhou, Mingxu
    and Li, Meng and Wang, Miaojun and Li, Mingming
    and Tian, Ning and Huang, Panpan and Zhang, Peng
    and Wang, Qiancheng and Chen, Qinyu and Du, Qiushi
    and Ge, Ruiqi and Zhang, Ruisong and Pan, Ruizhe
    and Wang, Runji and Chen, R. J. and Jin, R. L.
    and Chen, Ruyi and Lu, Shanghao and Zhou, Shangyan
    and Chen, Shanhuang and Ye, Shengfeng and Wang, Shiyu
    and Yu, Shuiping and Zhou, Shunfeng and Pan, Shuting
    and Li, S. S. and Zhou, Shuang and Wu, Shaoqing
    and Yun, Tao and Pei, Tian and Sun, Tianyu
    and Wang, T. and Zeng, Wangding and Liu, Wen
    and Liang, Wenfeng and Gao, Wenjun and Yu, Wenqin
    and Zhang, Wentao and Xiao, W. L. and An, Wei
    and Liu, Xiaodong and Wang, Xiaohan and Chen, Xiaokang
    and Nie, Xiaotao and Cheng, Xin and Liu, Xin
    and Xie, Xin and Liu, Xingchao and Yang, Xinyu
    and Li, Xinyuan and Su, Xuecheng and Lin, Xuheng
    and Li, X. Q. and Jin, Xiangyue and Shen, Xiaojin
    and Chen, Xiaosha and Sun, Xiaowen and Wang, Xiaoxiang
    and Song, Xinnan and Zhou, Xinyi and Wang, Xianzu
    and Shan, Xinxia and Li, Y. K. and Wang, Y. Q.
    and Wei, Y. X. and Zhang, Yang and Xu, Yanhong
    and Li, Yao and Zhao, Yao and Sun, Yaofeng
    and Wang, Yaohui and Yu, Yi and Zhang, Yichao
    and Shi, Yifan and Xiong, Yiliang and He, Ying
    and Piao, Yishi and Wang, Yisong and Tan, Yixuan
    and Ma, Yiyang and Liu, Yiyuan and Guo, Yongqiang
    and Ou, Yuan and Wang, Yuduan and Gong, Yue
    and Zou, Yuheng and He, Yujia and Xiong, Yunfan
    and Luo, Yuxiang and You, Yuxiang and Liu, Yuxuan
    and Zhou, Yuyang and Zhu, Y. X. and Huang, Yanping
    and Li, Yaohui and Zheng, Yi and Zhu, Yuchen
    and Ma, Yunxian and Tang, Ying and Zha, Yukun
    and Yan, Yuting and Ren, Z. Z. and Ren, Zehui
    and Sha, Zhangli and Fu, Zhe and Xu, Zhean
    and Xie, Zhenda and Zhang, Zhengyan and Hao, Zhewen
    and Ma, Zhicheng and Yan, Zhigang and Wu, Zhiyu
    and Gu, Zihui and Zhu, Zijia and Liu, Zijun
    and Li, Zilin and Xie, Ziwei and Song, Ziyang
    and Pan, Zizheng and Huang, Zhen and Xu, Zhipeng
    and Zhang, Zhongyu and Zhang, Zhen},
  title = {DeepSeek-R1 incentivizes reasoning in LLMs through reinforcement learning},
  journal = {Nature},
  year = {2025},
  month = sep,
  volume = {645},
  number = {8081},
  pages = {633--638},
  issn = {1476-4687},
  doi = {10.1038/s41586-025-09422-z},
}

@inproceedings{
yu2025dapo,
title={{DAPO}: An Open-Source {LLM} Reinforcement Learning System at Scale},
author={Qiying Yu and Zheng Zhang and Ruofei Zhu and Yufeng Yuan and Xiaochen Zuo and Yu Yue and Weinan Dai and Tiantian Fan and Gaohong Liu and Juncai Liu and LingJun Liu and Xin Liu and Haibin Lin and Zhiqi Lin and Bole Ma and Guangming Sheng and Yuxuan Tong and Chi Zhang and Mofan Zhang and Ru Zhang and Wang Zhang and Hang Zhu and Jinhua Zhu and Jiaze Chen and Jiangjie Chen and Chengyi Wang and Hongli Yu and Yuxuan Song and Xiangpeng Wei and Hao Zhou and Jingjing Liu and Wei-Ying Ma and Ya-Qin Zhang and Lin Yan and Yonghui Wu and Mingxuan Wang},
booktitle={The Thirty-ninth Annual Conference on Neural Information Processing Systems},
year={2025},
}

@article{bai2025qwen25vl,
  title         = {{Qwen2.5-VL} Technical Report},
  author        = {Bai, Shuai and Chen, Keqin and Liu, Xuejing and Wang, Jialin and Ge, Wenbin and Song, Sibo and Dang, Kai and Wang, Peng and Wang, Shijie and Tang, Jun and Zhong, Humen and Zhu, Yuanzhi and Yang, Mingkun and Li, Zhaohai and Wan, Jianqiang and Wang, Pengfei and Ding, Wei and Fu, Zheren and Xu, Yiheng and Ye, Jiabo and Zhang, Xi and Xie, Tianbao and Cheng, Zesen and Zhang, Hang and Yang, Zhibo and Xu, Haiyang and Lin, Junyang},
  journal       = {arXiv preprint arXiv:2502.13923},
  year          = {2025},
  eprint        = {2502.13923},
  archiveprefix = {arXiv},
}

@inproceedings{huang2025vppo,
 author = {Huang, Siyuan and Qu, Xiaoye and Li, Yafu and Luo, Yun and He, Zefeng and Liu, Daizong and Cheng, Yu},
 booktitle = {International Conference on Learning Representations},
 editor = {C. Vondrick and B. Hariharan and C. Raffel and L. Pinto and D. Yang and A. Faust},
 pages = {103291--103325},
 title = {Spotlight on Token Perception for Multimodal Reinforcement Learning},
 volume = {2026},
 year = {2026}
}

@article{yuan2026visionopd,
  title         = {{Vision-OPD}: Learning to See Fine Details for Multimodal {LLM}s via On-Policy Self-Distillation},
  author        = {Yuan, Qianhao and Lou, Jie and Yu, Xing and Lin, Hongyu and Sun, Le and Han, Xianpei and Lu, Yaojie},
  journal       = {arXiv preprint arXiv:2605.18740},
  year          = {2026},
  eprint        = {2605.18740},
  archiveprefix = {arXiv}
}

@article{liu2026vaopd,
  title         = {Visual-Advantage On-Policy Distillation for Vision-Language Models},
  author        = {Liu, Ruiqi and Lv, Xiaolei and Li, Gengsheng and Zhu, Ximo and Wang, Zhiheng and Zhang, Zhengbo and Chen, Junkai and Li, Zhiheng and Li, Bo and Gao, Jun and Wu, Shu},
  journal       = {arXiv preprint arXiv:2605.21924},
  year          = {2026},
  eprint        = {2605.21924},
  archiveprefix = {arXiv},
}

@article{zhang2026vad,
  title         = {{VAD}: Attributing Visual Evidence for Target Reconstruction in Multimodal On-Policy Distillation},
  author        = {Zhang, Kangning and Li, Yixing and Shao, Shuai and Li, Qingyao and Lu, Zhengxi and Yao, Zhiyuan and Lin, Jianghao and Jiao, Wenxiang and Lu, Yuan and Liu, Weiwen and Zhang, Weinan and Yu, Yong},
  journal       = {arXiv preprint arXiv:2607.28590},
  year          = {2026},
  eprint        = {2607.28590},
  archiveprefix = {arXiv},
}

@article{aniri2026opdv,
  title         = {{OPD-V}: Visual On-Policy Self-Distillation with Modality Balance},
  author        = {{Aniri} and Bi, Jinhe and Liao, Peng and Jin, Zengjie and Tresp, Volker and Shen, Fei and Ma, Yunpu and Chua, Tat-Seng},
  journal       = {arXiv preprint arXiv:2608.05131},
  year          = {2026},
  eprint        = {2608.05131},
  archiveprefix = {arXiv},
}

@article{sun2026vzero,
  title         = {{V-Zero}: Answer-Label-Free On-Policy Distillation with Contrastive Evidence Gating for Fine-Grained Visual Reasoning},
  author        = {Sun, Haoxiang and Yi, Zhihang and Deng, Langxuan and Zhou, Yuhao and Jia, Peiqi and Zhao, Jian and Yuan, Li and Lv, Jiancheng and Wang, Tao},
  journal       = {arXiv preprint arXiv:2606.25319},
  year          = {2026},
  eprint        = {2606.25319},
  archiveprefix = {arXiv},
}

@inproceedings{wang2025papo,
 author = {Wang, Zhenhailong and Guo, Xuehang and Stoica, Sofia and Xu, Haiyang and WANG, Hongru and Ha, Hyeonjeong and Chen, Xiusi and Chen, Yangyi and Yan, Ming and Huang, Fei and Ji, Heng},
 booktitle = {International Conference on Learning Representations},
 editor = {C. Vondrick and B. Hariharan and C. Raffel and L. Pinto and D. Yang and A. Faust},
 pages = {80502--80530},
 title = {Perception-Aware Policy Optimization for Multimodal Reasoning},
 volume = {2026},
 year = {2026}
}

@inproceedings{tian2025vapo,
 author = {Tian, Xinyu and Zou, Shu and Yang, Zhaoyuan and He, Mengqi and Waschkowski, Fabian and Wesemann, Lukas and Tu, Peter and Zhang, Jing},
 booktitle = {International Conference on Learning Representations},
 editor = {C. Vondrick and B. Hariharan and C. Raffel and L. Pinto and D. Yang and A. Faust},
 pages = {133928--133965},
 title = {More Thought, Less Accuracy? On the Dual Nature of Reasoning in Vision-Language Models},
 volume = {2026},
 year = {2026}
}

@inproceedings{wang2026vgpo,
    title = "Visually-Guided Policy Optimization for Multimodal Reasoning",
    author = "Wang, Zengbin  and
      Xiong, Feng  and
      Lin, Liang  and
      Hu, Xuecai  and
      Wang, Yong  and
      Wang, Yanlin  and
      Zhang, Man  and
      Chu, Xiangxiang",
    editor = "Liakata, Maria  and
      Moreira, Viviane P.  and
      Zhang, Jiajun  and
      Jurgens, David",
    booktitle = "Proceedings of the 64th Annual Meeting of the {A}ssociation for {C}omputational {L}inguistics (Volume 1: Long Papers)",
    month = jul,
    year = "2026",
    address = "San Diego, California, United States",
    publisher = "Association for Computational Linguistics",
    doi = "10.18653/v1/2026.acl-long.301",
    pages = "6646--6664",
    ISBN = "979-8-89176-390-6"
}

@inproceedings{liu2025noisyrollout,
 author = {Liu, Xiangyan and Ni, Jinjie and Wu, Zijian and Du, Chao and Dou, Longxu and Wang, Haonan and Pang, Tianyu and Shieh, Michael},
 booktitle = {Advances in Neural Information Processing Systems},
 doi = {10.52202/085713-0102},
 editor = {D. Belgrave and C. Zhang and H. Lin and R. Pascanu and P. Koniusz and M. Ghassemi and N. Chen},
 pages = {2923--2957},
 publisher = {Curran Associates, Inc.},
 title = {NoisyRollout: Reinforcing Visual Reasoning with Data Augmentation},
 volume = {38, Main Conference},
 year = {2025}
}

@inproceedings{wang2025vlrethinker,
 author = {Wang, Haozhe and Qu, Chao and Huang, Zuming and Chu, Wei and Lin, Fangzhen and Chen, Wenhu},
 booktitle = {Advances in Neural Information Processing Systems},
 doi = {10.52202/085713-1037},
 editor = {D. Belgrave and C. Zhang and H. Lin and R. Pascanu and P. Koniusz and M. Ghassemi and N. Chen},
 pages = {30865--30891},
 publisher = {Curran Associates, Inc.},
 title = {VL-Rethinker: Incentivizing Self-Reflection of Vision-Language Models with Reinforcement Learning},
 volume = {38, Main Conference},
 year = {2025}
}

@inproceedings{
ding2026octopus,
title={Learning Self-Correction in Vision{\textendash}Language Models via Rollout Augmentation},
author={Yi Ding and Ziliang Qiu and Bolian Li and Ruqi Zhang},
booktitle={Forty-third International Conference on Machine Learning},
year={2026}
}

@inproceedings{miao2026prco,
    title = "Seeing with You: Perception-Reasoning Co-evolution for Multimodal Reasoning",
    author = "Miao, Ziqi  and
      Jia, Haonan  and
      Li, Lijun  and
      Qian, Chen  and
      Xiong, Yuan  and
      Yan, Wenting  and
      Shao, Jing",
    editor = "Lee, Hung-Yi  and
      Dev, Sunipa  and
      Li, Junyi Jessy  and
      Elliott, Desmond",
    booktitle = "Proceedings of the 2026 Conference on Empirical Methods in Natural Language Processing",
    month = oct,
    year = "2026",
    address = "Budapest, Hungary",
    publisher = "Association for Computational Linguistics",
    doi = "10.18653/2026.emnlp-main.2363",
    pages = "53668--53690",
    ISBN = "979-8-89176-459-0"
}

@article{lu2026contextrl,
  title         = {{ContextRL}: Enhancing {MLLM}'s Knowledge Discovery Efficiency with Context-Augmented {RL}},
  author        = {Lu, Xingyu and Wang, Jinpeng and Zhang, Yifan and Ma, Shijie and Hu, Xiao and Zhang, Tianke and Fan, Haonan and Jiang, Kaiyu and Liu, Changyi and Tang, Kaiyu and Wen, Bin and Yang, Fan and Gao, Tingting and Li, Han and Yuan, Chun},
  journal       = {arXiv preprint arXiv:2602.22623},
  year          = {2026},
  eprint        = {2602.22623},
  archiveprefix = {arXiv}
}

@article{huang2026evidencerl,
  title         = {{Evidence-RL}: Towards Evidence-intensive Visual Reasoning},
  author        = {Huang, Haojie and Yu, Xinlei and Xu, Chengming and Chen, Zhangquan and Yang, Cheng and He, Qingdong and Yang, Yu and Zhang, Jiangning and Hu, Xiaobin},
  journal       = {arXiv preprint arXiv:2608.08021},
  year          = {2026},
  eprint        = {2608.08021},
  archiveprefix = {arXiv},
  primaryclass  = {cs.CV}
}

@inproceedings{xiao2025perceptionr1,
 author = {Xiao, Tong and Xu, Xin and Huang, Zhenya and Gao, Hongyu and Liu, Quan and Liu, Qi and Chen, Enhong},
 booktitle = {International Conference on Learning Representations},
 editor = {C. Vondrick and B. Hariharan and C. Raffel and L. Pinto and D. Yang and A. Faust},
 pages = {26868--26898},
 title = {Perception-R1: Advancing Multimodal Reasoning Capabilities of MLLMs via Visual Perception Reward},
 volume = {2026},
 year = {2026}
}

@inproceedings{gong2026saliencyr1,
  title     = {{Saliency-R1}: Enforcing Interpretable and Faithful Vision-Language Reasoning via Saliency-Map Alignment Reward},
  author    = {Gong, Shizhan and Hu, Minda and Zhang, Qiyuan and Ma, Chen and Dou, Qi},
  booktitle = {Proceedings of the IEEE/CVF Conference on Computer Vision and Pattern Recognition},
  year      = {2026}
}

@inproceedings{li2025tam,
  title     = {Token Activation Map to Visually Explain Multimodal LLMs},
  author    = {Li, Yi and Wang, Hualiang and Ding, Xinpeng and Wang, Haonan and Li, Xiaomeng},
  booktitle = {Proceedings of the IEEE/CVF International Conference on Computer Vision},
  year      = {2025},
  pages     = {48--58}
}

@inproceedings{yu2026cfpo,
  title     = {{CFPO}: Counterfactual Policy Optimization for Multimodal Reasoning},
  author    = {Yu, Zhangyuan and Sun, Wanran and Yang, Guangjing and Wu, Xiaohu and Lao, Qicheng},
  booktitle = {International Conference on Machine Learning},
  year      = {2026},
  eprint    = {2606.23206},
  archiveprefix = {arXiv}
}

@inproceedings{
liu2026seeing,
title={Seeing but Not Believing: Probing the Disconnect Between Visual Attention and Answer Correctness in {VLM}s},
author={Zhining Liu and Ziyi Chen and Hui Liu and Chen Luo and Xianfeng Tang and Suhang Wang and Jingying Zeng and Zhenwei Dai and Zhan Shi and Tianxin Wei and Hanqing Lu and Benoit Dumoulin and Hanghang Tong},
booktitle={The Fourteenth International Conference on Learning Representations},
year={2026}
}

@inproceedings{xiao-etal-2026-vl,
    title = "{VL}-Calibration: Decoupled Confidence Calibration for Large Vision-Language Models Reasoning",
    author = "Xiao, Wenyi  and
      Xu, Xinchi  and
      Gan, Leilei",
    editor = "Liakata, Maria  and
      Moreira, Viviane P.  and
      Zhang, Jiajun  and
      Jurgens, David",
    booktitle = "Proceedings of the 64th Annual Meeting of the {A}ssociation for {C}omputational {L}inguistics (Volume 1: Long Papers)",
    month = jul,
    year = "2026",
    address = "San Diego, California, United States",
    publisher = "Association for Computational Linguistics",
    doi = "10.18653/v1/2026.acl-long.2074",
    pages = "44791--44815",
    ISBN = "979-8-89176-390-6"
}

@inproceedings{xiao2025fastgrpo,
  author    = {Xiao, Wenyi and Gan, Leilei},
  title     = {Fast-Slow Thinking {GRPO} for Large Vision-Language Model Reasoning},
  booktitle = {Advances in Neural Information Processing Systems},
  volume    = {38},
  pages     = {190068--190098},
  year      = {2025},
  doi       = {10.52202/085713-5712}
}

@article{xiao2025hsadpo,
  author  = {Xiao, Wenyi and Huang, Ziwei and Gan, Leilei and He, Wanggui and
             Li, Haoyuan and Yu, Zhelun and Shu, Fangxun and Jiang, Hao and Zhu, Linchao},
  title   = {Detecting and Mitigating Hallucination in Large Vision Language Models via Fine-Grained {AI} Feedback},
  journal = {Proceedings of the {AAAI} Conference on Artificial Intelligence},
  volume  = {39},
  number  = {24},
  pages   = {25543--25551},
  year    = {2025},
  doi     = {10.1609/aaai.v39i24.34744}
}

@article{xiao2026dposurvey,
  author  = {Xiao, Wenyi and Wang, Zechuan and Gan, Leilei and Zhao, Shuai and
             Li, Zongyue and Lei, Ruirui and He, Wanggui and Tuan, Luu Anh and
             Chan, Long and Jiang, Hao and Zhao, Zhou and Wu, Fei},
  title   = {A Survey of Direct Preference Optimization: Datasets, Theories, Variants, and Applications},
  journal = {{IEEE} Transactions on Pattern Analysis and Machine Intelligence},
  volume  = {48},
  number  = {10},
  pages   = {13079--13098},
  year    = {2026},
  doi     = {10.1109/TPAMI.2026.3704314}
}

@inproceedings{shi2026revealer,
  author    = {Shi, Fulin and Xiao, Wenyi and Chen, Bin and Ding, Liang and Gan, Leilei},
  title     = {{REVEALER}: Reinforcement-Guided Visual Reasoning for Element-Level Text-Image Alignment Evaluation},
  booktitle = {Proceedings of the 64th Annual Meeting of the {A}ssociation for {C}omputational {L}inguistics (Volume 1: Long Papers)},
  pages     = {47630--47649},
  year      = {2026},
  publisher = {Association for Computational Linguistics},
  doi       = {10.18653/v1/2026.acl-long.2200}
}

@inproceedings{zhang2024mathverse,
  author = {Zhang, Renrui and Jiang, Dongzhi and Zhang, Yichi
            and Lin, Haokun and Guo, Ziyu and Qiu, Pengshuo
            and Zhou, Aojun and Lu, Pan and Chang, Kai-Wei
            and Qiao, Yu and Gao, Peng and Li, Hongsheng},
  editor = {Leonardis, Ale{\v{s}} and Ricci, Elisa and Roth, Stefan
            and Russakovsky, Olga and Sattler, Torsten and Varol, G{\"u}l},
  title = {MATHVERSE: Does Your Multi-modal LLM Truly See the Diagrams in Visual Math Problems?},
  booktitle = {Computer Vision -- ECCV 2024},
  year = {2025},
  publisher = {Springer Nature Switzerland},
  address = {Cham},
  pages = {169--186},
  isbn = {978-3-031-73242-3}
}

@inproceedings{lu2024mathvista,
  title     = {{MathVista}: Evaluating Mathematical Reasoning of Foundation Models in Visual Contexts},
  author    = {Lu, Pan and Bansal, Hritik and Xia, Tony and Liu, Jiacheng and Li, Chunyuan and Hajishirzi, Hannaneh and Cheng, Hao and Chang, Kai-Wei and Galley, Michel and Gao, Jianfeng},
  booktitle = {International Conference on Learning Representations},
  year      = {2024}
}

@inproceedings{zou2025dynamath,
  title     = {{DynaMath}: A Dynamic Visual Benchmark for Evaluating Mathematical Reasoning Robustness of Vision Language Models},
  author    = {Zou, Chengke and Guo, Xingang and Yang, Rui and Zhang, Junyu and Hu, Bin and Zhang, Huan},
  booktitle = {International Conference on Learning Representations},
  year      = {2025}
}

@inproceedings{wang2024mathvision,
  author = {Wang, Ke and Pan, Junting and Shi, Weikang and Lu, Zimu
            and Ren, Houxing and Zhou, Aojun and Zhan, Mingjie
            and Li, Hongsheng},
  title = {Measuring Multimodal Mathematical Reasoning with MATH-Vision Dataset},
  booktitle = {Advances in Neural Information Processing Systems},
  editor = {A. Globerson and L. Mackey and D. Belgrave and A. Fan
            and U. Paquet and J. Tomczak and C. Zhang},
  volume = {37},
  pages = {95095--95169},
  publisher = {Curran Associates, Inc.},
  year = {2024},
  doi = {10.52202/079017-3014},
}

@article{meng2025mmeureka,
  title = {{MM}-Eureka: Toward Stable Multimodal Reasoning via Rule-based Reinforcement Learning with Policy Drift Control},
  author = {Fanqing Meng and Lingxiao Du and Zongkai Liu and Zhixiang Zhou
            and Quanfeng Lu and Tiancheng Han and Daocheng Fu
            and Kaipeng Zhang and Ping Luo and Yu Qiao
            and Jiaheng Zhang and Michael Qizhe Shieh
            and Qiaosheng Zhang and Wenqi Shao},
  journal = {Transactions on Machine Learning Research},
  issn = {2835-8856},
  year = {2026}
}

@inproceedings{lu2022scienceqa,
  title     = {Learn to Explain: Multimodal Reasoning via Thought Chains for Science Question Answering},
  author    = {Lu, Pan and Mishra, Swaroop and Xia, Tony and Qiu, Liang and Chang, Kai-Wei and Zhu, Song-Chun and Tafjord, Oyvind and Clark, Peter and Kalyan, Ashwin},
  booktitle = {Advances in Neural Information Processing Systems},
  year      = {2022}
}

@inproceedings{qiao2025wemath,
  title     = {{We-Math}: Does Your Large Multimodal Model Achieve Human-Like Mathematical Reasoning?},
  author    = {Qiao, Runqi and Tan, Qiuna and Dong, Guanting and Wu, Minhui and Sun, Chong and Song, Xiaoshuai and Wang, Jiapeng and GongQue, Zhuoma and Lei, Shanglin and Zhang, Yifan and Wei, Zhe and Zhang, Miaoxuan and Qiao, Runfeng and Zong, Xiao and Xu, Yida and Yang, Peiqing and Bao, Zhimin and Diao, Muxi and Li, Chen and Zhang, Honggang},
  booktitle = {Proceedings of the 63rd Annual Meeting of the Association for Computational Linguistics},
  pages     = {20023--20070},
  year      = {2025}
}

@inproceedings{lu2021intergps,
  title     = {{Inter-GPS}: Interpretable Geometry Problem Solving with Formal Language and Symbolic Reasoning},
  author    = {Lu, Pan and Gong, Ran and Jiang, Shibiao and Qiu, Liang and Huang, Siyuan and Liang, Xiaodan and Zhu, Song-Chun},
  booktitle = {Proceedings of the 59th Annual Meeting of the Association for Computational Linguistics and the 11th International Joint Conference on Natural Language Processing},
  pages     = {6774--6786},
  year      = {2021}
}

@article{xiao2024logicvista,
  title         = {{LogicVista}: Multimodal {LLM} Logical Reasoning Benchmark in Visual Contexts},
  author        = {Xiao, Yijia and Sun, Edward and Liu, Tianyu and Wang, Wei},
  journal       = {arXiv preprint arXiv:2407.04973},
  year          = {2024},
  eprint        = {2407.04973},
  archiveprefix = {arXiv}
}

@inproceedings{johnson2017clevr,
  title     = {{CLEVR}: A Diagnostic Dataset for Compositional Language and Elementary Visual Reasoning},
  author    = {Johnson, Justin and Hariharan, Bharath and van der Maaten, Laurens and Fei-Fei, Li and Zitnick, C. Lawrence and Girshick, Ross},
  booktitle = {Proceedings of the IEEE Conference on Computer Vision and Pattern Recognition},
  pages     = {2901--2910},
  year      = {2017}
}

@inproceedings{guan2024hallusionbench,
  title     = {{HallusionBench}: An Advanced Diagnostic Suite for Entangled Language Hallucination and Visual Illusion in Large Vision-Language Models},
  author    = {Guan, Tianrui and Liu, Fuxiao and Wu, Xiyang and Xian, Ruiqi and Li, Zongxia and Liu, Xiaoyu and Wang, Xijun and Chen, Lichang and Huang, Furong and Yacoob, Yaser and Manocha, Dinesh and Zhou, Tianyi},
  booktitle = {Proceedings of the IEEE/CVF Conference on Computer Vision and Pattern Recognition},
  pages     = {14375--14385},
  year      = {2024}
}

@inproceedings{rahmanzadehgervi2024blind,
  title     = {Vision Language Models Are Blind},
  author    = {Rahmanzadehgervi, Pooyan and Bolton, Logan and Taesiri, Mohammad Reza and Nguyen, Anh Totti},
  booktitle = {Proceedings of the Asian Conference on Computer Vision},
  pages     = {18--34},
  year      = {2024}
}

@inproceedings{wu2024vstar,
  title     = {{V*}: Guided Visual Search as a Core Mechanism in Multimodal {LLM}s},
  author    = {Wu, Penghao and Xie, Saining},
  booktitle = {Proceedings of the IEEE/CVF Conference on Computer Vision and Pattern Recognition},
  pages     = {13084--13094},
  year      = {2024},
}

@inproceedings{yangCVPR2025visionzip,
  title={Visionzip: Longer is better but not necessary in vision language models},
  author={Yang, Senqiao and Chen, Yukang and Tian, Zhuotao and Wang, Chengyao and Li, Jingyao and Yu, Bei and Jia, Jiaya},
  booktitle={2025 IEEE/CVF Conference on Computer Vision and Pattern Recognition (CVPR)},
  pages={19792--19802},
  year={2025},
  organization={IEEE}
}

@inproceedings{yue2025mmmupro,
  title = {{MMMU}-Pro: A More Robust Multi-discipline Multimodal Understanding Benchmark},
  author = {Yue, Xiang and Zheng, Tianyu and Ni, Yuansheng and Wang, Yubo and Zhang, Kai and Tong, Shengbang and Sun, Yuxuan and Yu, Botao and Zhang, Ge and Sun, Huan and Su, Yu and Chen, Wenhu and Neubig, Graham},
  booktitle = {Proceedings of the 63rd Annual Meeting of the Association for Computational Linguistics (Volume 1: Long Papers)},
  pages = {15134--15186},
  year = {2025},
  doi = {10.18653/v1/2025.acl-long.736}
}

@inproceedings{chen2024mmstar,
  title = {Are We on the Right Way for Evaluating Large Vision-Language Models?},
  author = {Chen, Lin and Li, Jinsong and Dong, Xiaoyi and Zhang, Pan and Zang, Yuhang and Chen, Zehui and Duan, Haodong and Wang, Jiaqi and Qiao, Yu and Lin, Dahua and Zhao, Feng},
  booktitle = {Advances in Neural Information Processing Systems},
  volume = {37},
  pages = {27056--27087},
  year = {2024},
  doi = {10.52202/079017-0850}
}

@inproceedings{tong2024cambrian,
  title = {Cambrian-1: A Fully Open, Vision-Centric Exploration of Multimodal {LLMs}},
  author = {Tong, Shengbang and Brown, Ellis and Wu, Penghao and Woo, Sanghyun and Middepogu, Manoj and Akula, Sai Charitha and Yang, Jihan and Yang, Shusheng and Iyer, Adithya and Pan, Xichen and Wang, Austin and Fergus, Rob and LeCun, Yann and Xie, Saining},
  booktitle = {Advances in Neural Information Processing Systems},
  volume = {37},
  pages = {87310--87356},
  year = {2024},
  doi = {10.52202/079017-2771}
}

\clearpage
\appendix
\section*{Appendix Contents}
\begingroup
\makeatletter
\newcommand{\appendixsectionentry}[2]{%
  \@dottedtocline{1}{0em}{1.5em}{\hyperref[#1]{\numberline{\ref*{#1}}#2}}{\pageref*{#1}}%
}
\newcommand{\appendixsubsectionentry}[2]{%
  \@dottedtocline{2}{1.5em}{2.3em}{\hyperref[#1]{\numberline{\ref*{#1}}#2}}{\pageref*{#1}}%
}
\appendixsectionentry{app:limitations}{Limitations}
\appendixsectionentry{app:experimental_setup}{Experimental Setup}
\appendixsubsectionentry{app:training_details}{Training Details}
\appendixsubsectionentry{app:evaluation_details}{Evaluation Details}
\appendixsubsectionentry{app:qwen25_ablation}{Ablation Studies}
\appendixsectionentry{app:pilot_details}{Pilot Study Details}
\appendixsubsectionentry{app:pilot1_details}{VEPs Distinguish Correct and
Incorrect Rollouts}
\appendixsubsectionentry{app:pilot2_details}{Contrastive VEPs Identify Supporting Visual Tokens}
\appendixsubsectionentry{app:pilot3_details}{Contrastive VEPs Locate Supporting and Harmful Response Tokens}
\appendixsectionentry{app:theory}{Theoretical Analysis}
\appendixsubsectionentry{app:prototype_recovery}{Contrastive Prototype Recovery}
\appendixsubsectionentry{app:local_improvement}{Local Policy Improvement}
\appendixsectionentry{app:additional_analysis}{Additional Analysis}
\appendixsubsectionentry{app:error_analysis}{Error Analysis}
\appendixsubsectionentry{app:token_frequencies}{Selected-token Frequencies}
\appendixsubsectionentry{app:case_study}{Case Study}
\makeatother
\endgroup
\bigskip
\section{Limitations}
\label{app:limitations}

While our results demonstrate the effectiveness of VEP and VEP-RL, it is important to acknowledge their current
limitations and future directions.

\textbf{Computational Overhead.} Our method introduces a limited computational overhead. To construct VEPs and compute token-level guidance, VEP-RL requires an additional intervened forward pass for rollouts in eligible groups. Empirically, we found this increases total training time by approximately 3--4\%
relative to GRPO (e.g., from 58.3 to 60.2 hours
on Qwen2.5-VL-7B). 

\textbf{Scope of Generalization.} Our experiments have demonstrated the effectiveness of VEP-RL across three models. We have not yet verified the efficacy of our method on extremely large-scale models (e.g., 32B+ parameters) due to computational resource constraints.

\section{Experimental Setup}
\label{app:experimental_setup}

\subsection{Training Details}
\label{app:training_details}

\begin{table}[H]
    \centering
    \small
    \setlength{\tabcolsep}{5pt}
    \caption{Training hyperparameters used for GRPO, DAPO, and VEP-RL.}
    \label{tab:training_hyperparameters}
    \begin{tabular}{@{}p{0.47\columnwidth}p{0.45\columnwidth}@{}}
        \toprule
        \textbf{Hyperparameter} & \textbf{Value} \\
        \midrule
        Training data & ViRL-39K \\
        Training epochs & 2 \\
        Optimizer & AdamW \\
        Learning rate & \(1\times10^{-6}\) \\
        Actor global batch size & 128 \\
        Rollout batch size & 256 \\
        Group size \(G\) & 8 \\
        Rollout temperature / top-\(p\) & 1.0 / 1.0 \\
        Training data order & Shuffled with seed 42 \\
        Rollout sampling seed & 42 \\
        Maximum prompt / response length & 4096 / 4096 \\
        Precision & BF16 \\
        Minimum / maximum image pixels & 262,144 / 4,000,000 \\
        \revision{GRPO / VEP-RL KL coefficient} & \revision{\(1\times10^{-4}\)} \\
        \revision{DAPO KL regularization} & \revision{Disabled} \\
        \revision{GRPO / VEP-RL clip-low / clip-high} & \revision{0.20 / 0.20} \\
        \revision{DAPO clip-low / clip-high} & \revision{0.20 / 0.28} \\
        \revision{Dynamic sampling / overlong reward shaping} & \revision{DAPO only} \\
        Verifier & MathRuler boxed-answer grader \\
        Accuracy / format reward weight & 0.9 / 0.1 \\
        VEP visual-token selection ratio & 10\% \\
        VEP response-token selection & Top 5\% positive scores \\
        VEP \(\lambda_{\mathrm{traj}} / \lambda_{\mathrm{tok}}\) & 0.1 / 0.1 \\
        \bottomrule
    \end{tabular}
\end{table}

\revision{For comparison with VPPO~\citep{huang2025vppo} and
PAPO~\citep{wang2025papo}, we train GRPO, DAPO, and VEP-RL on ViRL-39K for two
epochs from Qwen2.5-VL-7B-Instruct, Qwen3-VL-8B-Instruct, and Qwen3.5-9B.
These runs follow VPPO in training data and number of epochs. Method-specific settings are listed in
Table~\ref{tab:training_hyperparameters}.}

\paragraph{GRPO objective.}
Using the
importance ratio defined in the main text, our GRPO objective is
\begin{equation}
\mathcal{J}_{\mathrm{GRPO}}(\theta)
=
\mathbb{E}_{Q,\{r_i\}}
\left[
\sum_{i=1}^{G}\sum_{t=1}^{|r_i|}
\frac{
\min\!\left\{
\rho_{i,t}\hat A_i,
\operatorname{clip}\!\left(
\rho_{i,t},1-\varepsilon_{\mathrm{low}},
1+\varepsilon_{\mathrm{high}}
\right)\hat A_i
\right\}
-\beta D_{\mathrm{KL}}^{i,t}
}{G|r_i|}
\right].
\label{eq:grpo_objective}
\end{equation}
Here, $\hat A_i$ is the group-normalized trajectory advantage
computed from the rollout reward. The clipping bounds and KL
regularization are shared with VEP-RL.

\paragraph{General Training Hyperparameters.}
We use full-parameter training for two epochs, including the vision tower,
with AdamW, a learning rate of \(10^{-6}\), and BF16 precision. Each question
produces eight rollouts sampled with temperature \(1.0\) and top-\(p=1.0\).
The rollout batch size is 256 and the actor global batch size is 128. Both the
prompt and response lengths are capped at 4096 tokens. Images contain between
262,144 and 4,000,000 pixels after preprocessing.
\revision{GRPO and VEP-RL use symmetric clipping with clip-low and clip-high
both set to 0.20, together with low-variance KL regularization with coefficient
\(10^{-4}\). VEP-RL retains the GRPO recipe and adds only trajectory- and
token-level advantage reweighting. DAPO uses clip-low 0.20 and clip-high 0.28,
dynamic sampling, and overlong reward shaping, with KL regularization disabled.}
We shuffle the
training data with seed 42 and use the same seed for rollout sampling.

Each prompt requires the reasoning process inside
\texttt{<think>...</think>} and the final answer inside
\(\backslash\texttt{boxed\{...\}}\). We use MathRuler to extract the final
boxed content and compare it with the reference answer. The reward assigns
weights 0.9 and 0.1 to answer correctness and format compliance,
respectively; other answer-like spans are not considered.

\paragraph{VEP-RL-specific Hyperparameters.}
We average attention over heads and over the decoder layers at 45\%, 65\%,
and 85\% of model depth, rounding each index upward. The resulting one-based
layer indices are listed in Table~\ref{tab:vep_decoder_layers}. For each
question, the Success and Failure VEPs
are the mean attributions of its correct and incorrect rollouts. We rank visual
tokens directly by the signed difference between these prototypes, without
taking the positive part or renormalizing the scores. We apply VEP guidance
only when a rollout group contains at least two correct and two incorrect
rollouts. Otherwise, we use the original update in
Eq.~\eqref{eq:grpo_objective}.

For token-level guidance, correct rollouts use \(\mathbf D_Q^+\) and incorrect
rollouts use \(\mathbf D_Q^-\). We replace the top 10\% selected visual states
with the mean visual state of the same image before the first language-decoder
layer. For Qwen3-VL, we apply the same replacement to each DeepStack visual
stream. We record the original-image sampled-token log probabilities during
rollout generation. For groups eligible for VEP guidance, we teacher-force the
sampled response once with the selected visual states replaced to obtain the
intervened log probabilities. Candidate positions are all response tokens. We rank them by \(\Delta_{r,t}\) from
Eq.~\eqref{eq:token_logp_change} and select up to the top 5\%, retaining only
positions with positive scores. Final-answer tokens remain eligible during
training.

Prototype construction and token selection use stop-gradient. If a rollout
has no eligible positive score, every response token keeps the trajectory-level
advantage. We set \(\lambda_{\mathrm{traj}}=\lambda_{\mathrm{tok}}=0.1\).

\begin{table}[H]
    \centering
    \small
    \setlength{\tabcolsep}{4pt}
    \caption{Decoder layers used to construct VEPs. Layer indices are
    one-based and are obtained by applying \(\lceil rL\rceil\) at depth ratios
    \(r\in\{0.45,0.65,0.85\}\). For Qwen3.5, we use the nearest full-attention layers at these depths.}
    \label{tab:vep_decoder_layers}
    \begin{tabular}{@{}lcc@{}}
        \toprule
        \textbf{Backbone} & \textbf{Depth \(L\)} & \textbf{Decoder layers} \\
        \midrule
        Qwen2.5-VL-7B & 28 & 13, 19, 24 \\
        Qwen3-VL-8B & 36 & 17, 24, 31 \\
        Qwen3.5-9B & 32 & 16, 20, 28 \\
        \bottomrule
    \end{tabular}
\end{table}

\paragraph{Computation Environment.}
Training uses one node with eight NVIDIA A800-SXM4 80GB GPUs. The
implementation extends VeRL and uses vLLM for rollout generation and
evaluation. We will release the training and evaluation code, configurations,
and model checkpoints.

\paragraph{Training Time.}
VPPO uses a masked-image forward pass for every rollout~\citep{huang2025vppo}.
VEP-RL uses an intervened pass only in groups with at least two correct and two
incorrect rollouts. Table~\ref{tab:training_time} compares training time and
throughput. The VEP-RL measurements include attention extraction and VEP
construction.

\begin{table}[H]
    \centering
    \caption{
Training time and throughput. VPPO rows are estimates obtained by applying
    its reported 9.7\% overhead~\citep{huang2025vppo} to the GRPO
    rows. Throughput is measured in questions per second, and overhead is
    relative to GRPO. Active (\%) is the fraction of training rollout groups
    with at least two correct and two incorrect responses.
}
    \label{tab:training_time}
    \begin{tabular}{@{}llcccc@{}}
        \toprule
        \textbf{Backbone} & \textbf{Method} & \textbf{Time (h)}
        & \textbf{Q/s} & \textbf{Overhead (\%)} & \textbf{Active (\%)} \\
        \midrule
        Qwen2.5-VL-7B & GRPO   & 58.3 & 0.370 & 0.00  & --   \\
                      & VEP-RL & 60.2 & 0.359 & +3.26 & 27.8 \\
                      & VPPO (est.) & 64.0 & 0.337 & +9.70 & --   \\
        \addlinespace
        Qwen3-VL-8B   & GRPO   & 75.8 & 0.285 & 0.00  & --   \\
                      & VEP-RL & 78.6 & 0.275 & +3.69 & 20.6 \\
                      & VPPO (est.) & 83.2 & 0.260 & +9.70 & --   \\
        \addlinespace
        Qwen3.5-9B    & GRPO   & 86.1 & 0.251 & 0.00  & --   \\
                      & VEP-RL & 89.2 & 0.242 & +3.60 & 17.9 \\
                      & VPPO (est.) & 94.5 & 0.229 & +9.70 & --   \\
        \bottomrule
    \end{tabular}
\end{table}

VEP-RL adds about 3.5\% training time over GRPO, compared with the 9.7\% VPPO
estimate according to its paper. Its intervened pass is used in 17.9--27.8\% of groups.

\subsection{Evaluation Details}
\label{app:evaluation_details}

\subsubsection{Models and Evaluation Settings}
\label{app:baseline_sources}

To ensure a fair comparison with the public baselines, we evaluate all models
on the complete benchmark files with the same scoring rules. For the public
baselines in Table~\ref{tab:main_results}, we follow the inference settings
in their official repositories. The training data and inference settings for
each model are listed below. For our trained GRPO, DAPO, and VEP-RL models on
all three backbones, we report average accuracy@8: we score all eight responses
to each question separately and average their correctness scores.
\revision{Thinking mode is disabled for all evaluated methods on Qwen3-VL-8B
and Qwen3.5-9B, including VEP-RL.}

\paragraph{Qwen2.5-VL-7B.}
\begin{itemize}
    \item \textbf{NoisyRollout}~\citep{liu2025noisyrollout}: Qwen2.5-VL-7B-Instruct, trained on Geo3K. Its evaluation script uses temperature 0, top-\(p=0.95\), one response, and a 2048-token response limit.
    \item \textbf{VL-Rethinker}~\citep{wang2025vlrethinker}: Qwen2.5-VL-7B-Instruct, trained on ViRL-39K. Its evaluation script uses temperature 1.0, top-\(p=0.95\), one response, and an 8192-token response limit.
    \item \textbf{PAPO}~\citep{wang2025papo}: Qwen2.5-VL-7B-Instruct, trained on ViRL-39K. PAPO-Eval uses temperature 1.0, top-\(p=1.0\), eight responses per question, and a 2048-token response limit.
    \item \textbf{VPPO}~\citep{huang2025vppo}: Qwen2.5-VL-7B-Instruct, trained on ViRL-39K. Its PAPO-Eval setting uses temperature 1.0, top-\(p=1.0\), eight responses per question, and a 2048-token response limit.
    \item \textbf{GRPO, DAPO, and VEP-RL}: Qwen2.5-VL-7B-Instruct, trained on ViRL-39K with the settings in Table~\ref{tab:training_hyperparameters}. Inference matches VPPO: temperature 1.0, top-\(p=1.0\), eight responses per question, and a 2048-token response limit.
    \item \textbf{VAPO}~\citep{tian2025vapo}: Qwen2.5-VL-7B-Instruct, trained on the ViRL-39K-derived VAPO-Thinker-36K set with visual anchors. Its VLMEvalKit/LMDeploy evaluation uses temperature 0, one response, and a 1024-token default response limit.
    \item \textbf{VGPO}~\citep{wang2026vgpo}: Qwen2.5-VL-7B-Instruct, trained on ViRL-39K. Its Look-Back-based evaluation uses temperature 0 and one response per question.
\end{itemize}

\paragraph{Qwen3-VL-8B.}
\begin{itemize}
    \item \textbf{OPD-V}~\citep{aniri2026opdv}: Qwen3-VL-8B-Instruct, trained on Vision-OPD-6K. Its evaluation script uses temperature 0, a 32,768-token response limit, and seed 42.
    \item \textbf{ContextRL}~\citep{lu2026contextrl}: Qwen3-VL-8B-Instruct, trained on a 29K-example set of VQA and multimodal math questions selected from FineVision. The paper uses VLMEvalKit for evaluation but does not report its decoding parameters; the 16K response limit is a training setting.
    \item \textbf{PRCO}~\citep{miao2026prco}: Qwen3-VL-8B-Instruct, trained on ViRL-39K. Its two-stage VLMEvalKit inference uses temperature 0 and top-\(p=1.0\) for both stages, with 1024 tokens for the Observer and 2048 for the Solver.
    \item \textbf{Octopus}~\citep{ding2026octopus}: Qwen3-VL-8B-Instruct, first trained on a 10K-example self-correction set and then on ViRL-39K. Its vLLM inference uses the self-correction prompt, temperature 1.0, top-\(p=0.95\), and a 16,384-token response limit.
    \item \textbf{VPPO}~\citep{huang2025vppo}: Qwen3-VL-8B-Instruct, trained on ViRL-39K. Its PAPO-Eval setting uses temperature 1.0, top-\(p=1.0\), eight responses per question, and a 2048-token response limit.
    \item \textbf{GRPO, DAPO, and VEP-RL}: Qwen3-VL-8B-Instruct, trained on ViRL-39K with the settings in Table~\ref{tab:training_hyperparameters}. Inference matches VPPO: temperature 1.0, top-\(p=1.0\), eight responses per question, and a 2048-token response limit.
\end{itemize}

\paragraph{Qwen3.5-9B.}
\begin{itemize}
    \item \textbf{OPD-V}~\citep{aniri2026opdv}: Qwen3.5-9B, trained on Vision-OPD-6K. Its evaluation script uses temperature 0, a 32,768-token response limit, and seed 42.
    \item \textbf{Vision-OPD}~\citep{yuan2026visionopd}: Qwen3.5-9B, trained on Vision-OPD-6K. Its evaluation script uses temperature 0, a 32,768-token response limit, and seed 42.
    \item \textbf{VAD}~\citep{zhang2026vad}: Qwen3.5-9B, trained on three-view examples built from Vision-OPD-6K. Its evaluation script uses temperature 0, an 8192-token response limit, and seed 42.
    \item \textbf{GRPO, DAPO, and VEP-RL}: Qwen3.5-9B, trained on ViRL-39K with the settings in Table~\ref{tab:training_hyperparameters}. Inference uses \revision{temperature 1.0, top-\(p=1.0\)}, eight responses per question, and an 8192-token response limit.
\end{itemize}

We report accuracy for each benchmark. Before scoring HallusionBench, we map
\emph{Yes} to 1 and \emph{No} to 0. For V*Bench, we report micro accuracy over
its 191 examples. For VLMsAreBlind, we use the official eight-group macro
average, keeping Olympic rings and pentagons separate and combining the word
and blank variants of the grid task. For CV-Bench, we average the ADE20K/COCO
2D score and the Omni3D 3D score. The category and overall scores in
Table~\ref{tab:main_results} are unweighted macro-averages of the reported
benchmark scores.

\subsubsection{Stability across Training Seeds}
\label{app:statistical_analysis}

We repeat DAPO and VEP-RL on Qwen3.5-9B with seeds 42, 43, and 44. The runs use
the same base checkpoint, training data, number of updates, batch sizes, and
sampling hyperparameters; the seed controls data order and rollout sampling.
Table~\ref{tab:qwen35_multiseed} reports the 14-benchmark macro accuracy for
every run.

\begin{table}[H]
    \centering
    \caption{Qwen3.5-9B results across three training seeds (14-benchmark
    macro accuracy, \%).}
    \label{tab:qwen35_multiseed}
    \small
    \setlength{\tabcolsep}{4.5pt}
    \begin{tabular}{@{}lcccc@{}}
        \toprule
        \textbf{Method} & \textbf{Seed 42} & \textbf{Seed 43}
        & \textbf{Seed 44} & \textbf{Mean $\pm$ std} \\
        \midrule
        DAPO & 84.64 & 84.18 & 84.79 & 84.54 $\pm$ 0.32 \\
        VEP-RL & 87.88 & 87.88 & 87.29 & 87.68 $\pm$ 0.34 \\
        VEP-RL $-$ DAPO & +3.24 & +3.70 & +2.50 & +3.15 $\pm$ 0.61 \\
        \bottomrule
    \end{tabular}
\end{table}

VEP-RL improves DAPO by $3.24$, $3.70$, and $2.50$ points across the three
training seeds, for a mean paired gain of $3.15 \pm 0.61$ points. Treating each
seed pair as one observation on the fixed 14-benchmark suite, the gain is
significant under a two-sided paired $t$-test
($t(2)=9.00$, $p=0.012$), with a 95\% confidence interval of $[1.64,4.65]$
points for the mean gain.

\subsubsection{Evaluation Benchmarks}
\label{app:evaluation_benchmarks}

We evaluate fourteen benchmarks covering mathematical reasoning, scientific
and logical reasoning, and vision-centric tasks.

\paragraph{Mathematical Reasoning.}
MathVerse varies how information is divided between the question and the
diagram, while MathVista places mathematical problems in charts, diagrams, and
synthetic scenes~\citep{zhang2024mathverse,lu2024mathvista}. MATH-Vision adds
competition-level problems across a broad range of subjects and difficulty
levels~\citep{wang2024mathvision}. DynaMath measures robustness to visual and
numerical variations of the same problem~\citep{zou2025dynamath}. We-Math
organizes problems by knowledge concept and reasoning step, and Geometry3K
focuses on geometry problems that require the diagram and text to be
interpreted together~\citep{qiao2025wemath,lu2021intergps}.

\paragraph{Scientific and Logical Reasoning.}
MMK12 contains K--12 mathematics and science problems with human-verified
answers~\citep{meng2025mmeureka}. ScienceQA covers questions
from natural, social, and language science~\citep{lu2022scienceqa}, while
LogicVista examines inductive, deductive, numerical, spatial, and mechanical
reasoning in visual contexts~\citep{xiao2024logicvista}.

\paragraph{Vision-centric Reasoning.}
MathVerse-V contains the vision-dependent MathVerse problems whose solutions
require information from the diagram~\citep{zhang2024mathverse}. CLEVR
Counting isolates object counting in controlled synthetic scenes
~\citep{johnson2017clevr}. HallusionBench studies visual illusion and language
hallucination~\citep{guan2024hallusionbench}, and VLMsAreBlind probes basic
visual operations such as counting, recognizing intersections, and judging
spatial relations~\citep{rahmanzadehgervi2024blind}. V*Bench tests fine-grained
visual search in high-resolution and crowded images~\citep{wu2024vstar}.

\paragraph{Out-of-domain Evaluation.}
For out-of-domain evaluation, we test Qwen3-VL and Qwen3.5 models on MMMU-Pro,
MM-Star, and CV-Bench. These benchmarks are not used for training. They test
multi-disciplinary reasoning, fine-grained perception, and 2D/3D spatial
understanding. We report the unweighted mean of the three benchmark scores.

\subsection{Ablation Studies}
\label{app:qwen25_ablation}

To give a further validation of VEP-RL, we additionally ablate trajectory- and token-level guidance on
Qwen2.5-VL-7B. Table~\ref{tab:qwen25_guidance_ablation} reports the two
category averages used in the main ablation study. Beyond ablation of components, we also test whether the gains could arise from random reweighting rather than VEP-based
selection.

\begin{table}[H]
    \centering
    \caption{VEP-RL guidance ablations on Qwen2.5-VL-7B. VR and VC denote
    Visual Reasoning and Vision-Centric category-average accuracy (\%).}
    \label{tab:qwen25_guidance_ablation}
    \small
    \setlength{\tabcolsep}{8pt}
    \begin{tabular}{@{}lcccc@{}}
        \toprule
        \textbf{Variant} & \textbf{Trajectory} & \textbf{Token}
        & \textbf{VR} & \textbf{VC} \\
        \midrule
        DAPO & -- & -- & 62.66 & 70.02 \\
        Trajectory only & $\checkmark$ & -- & 64.58 & 73.64 \\
        Token Shuffle & $\checkmark$ & Shuffle & 64.63 & 73.58 \\
        Token only & -- & $\checkmark$ & 64.13 & 72.87 \\
        Trajectory Shuffle & Shuffle & $\checkmark$ & 64.17 & 72.95 \\

        Excluded Final-Answer Token  & $\checkmark$ & No Final-Answer & 65.38 & 74.56 \\

        \textbf{VEP-RL} & $\checkmark$ & $\checkmark$
        & \textbf{65.64} & \textbf{74.90} \\
        \bottomrule
    \end{tabular}
\end{table}

First, shuffling trajectory weights while retaining token-level
guidance yields results close to the token-only variant (64.17/72.95
versus 64.13/72.87 on VR/VC). 
Similarly, shuffling token weights while
retaining trajectory-level guidance yields results close to the
trajectory-only variant (64.63/73.58 versus 64.58/73.64). 
Both are below
full VEP-RL (65.64/74.90), suggesting that where the weights are assigned
matters. 
Second, excluding final-answer tokens from token-level guidance retains
65.38/74.56, indicating that non-final-answer response tokens provide the main 
contribution to the improvement.
Third, trajectory- and token-level guidance each improve over DAPO in both
categories. Combining them gives the highest VR and VC accuracy.

\clearpage
\section{Pilot Study Details}
\label{app:pilot_details}

\paragraph{Setting.}
All three pilots use questions from the ViRL-39K training set and
evaluate Qwen2.5-VL, Qwen3-VL, and Qwen3.5.
Within each pilot, the compared methods use the same questions,
prompts, decoding settings, and sampling seeds.
We sample eight rollouts or continuations per condition and
compute confidence intervals over questions rather than
individual responses.
Selector comparisons use 20,000 paired question-level sign flips
with Holm correction.

\paragraph{Baselines.}
We compare rollout-level scores in Pilot~1, visual-token
selectors in Pilot~2, and response-token selectors in Pilot~3.
The following descriptions specify the scoring or selection
signals used in each pilot.

\noindent\textbf{Pilot~1: Distinguishing rollout outcomes.}
Pilot~1 compares the VEP Similarity Score with the Success
VEP Similarity, Answer Attention, Reasoning Attention,
and Shuffled VEP.
Prototypes are all constructed from a support set and evaluated
on independently sampled rollouts.

\begin{itemize}
    \item \textbf{VEP Similarity Score.}
    We construct the Success and Failure VEPs,
    $\boldsymbol{\mu}_Q^{+}$ and $\boldsymbol{\mu}_Q^{-}$.
    For an evaluation rollout $r$, we compute
    \[
        s_r =
        \cos(\mathbf{v}_r,\boldsymbol{\mu}_Q^{+})
        -
        \cos(\mathbf{v}_r,\boldsymbol{\mu}_Q^{-}),
    \]
    where $\mathbf{v}_r$ is its visual attribution.
    A larger score indicates greater similarity to the success prototype relative to the failure prototype.

    \item \textbf{Success VEP Similarity.}
    We use only similarity to the success prototype,
    $\cos(\mathbf{v}_r,\boldsymbol{\mu}_Q^{+})$.

    \item \textbf{Answer Attention.}
    We sum the direct attention assigned to visual tokens at each final-answer prediction position and average over these positions:
    \[
        s_r^{\mathrm{ans}}
        =
        \frac{1}{|\mathcal{A}_r|}
        \sum_{a \in \mathcal{A}_r}
        \sum_{i \in I_Q} A_r(a,i),
    \]
    where $\mathcal{A}_r$ and $I_Q$ denote final-answer token positions and visual positions, respectively. This score measures overall visual attention during final-answer prediction without outcome grouping.

    \item \textbf{Reasoning Attention.}
    We sum the direct attention assigned to visual tokens at each response prediction position and average over all response positions, including final-answer tokens:
    \[
        s_r^{\mathrm{attn}}
        =
        \frac{1}{|\mathcal{P}_r|}
        \sum_{t \in \mathcal{P}_r}
        \sum_{i \in I_Q} A_r(t,i),
    \]
    where $\mathcal{P}_r$ denotes all response token positions, including final-answer tokens, and $I_Q$ denotes visual positions.
    This score measures overall visual attention across the full response without outcome grouping.

    \item \textbf{Shuffled VEP.}
    We permute the support outcomes before constructing
    the Success and Failure VEPs, then compute the same
    two-prototype similarity score.
\end{itemize}

\noindent\textbf{Pilot~2: Selecting visual tokens.}
Pilot~2 compares seven visual-token selectors.
Each selector chooses 10\% of the visual states, which are replaced with the mean visual state of the same image.
We then sample eight complete responses per question and report the accuracy decrease relative to unintervened generation.
All selectors use the same intervention position and budget.

\begin{itemize}
    \item \textbf{Success-Specific Contrastive VEP.}
    We rank visual tokens by
    $\mathbf{D}_Q^{+}
    = \boldsymbol{\mu}_Q^{+}-\boldsymbol{\mu}_Q^{-}$ and select the highest-scoring positions.
    This emphasizes visual evidence attributed more strongly to successful rollouts than to failed rollouts.

    \item \textbf{Success VEP.}
    We rank visual tokens directly by $\boldsymbol{\mu}_Q^{+}$, without subtracting the failure prototype.
    This retains visual evidence shared by successful and failed rollouts.

    \item \textbf{Saliency-R1.}
    We use the saliency extraction method of Saliency-R1~\citep{gong2026saliencyr1}. It composes answer-to-reasoning and reasoning-to-visual attention weights to trace visual information through intermediate reasoning tokens.
    These weights are combined with visual value vectors, attention output projections, and target-token unembedding weights to estimate visual contributions to answer logits.
    The contributions are summed across layers and attention heads, followed by ReLU to retain positive contributions.
    The per-answer-token maps are then summed, without additional column normalization of the attention-rollout matrices, to obtain the final saliency map for ranking visual tokens.

    \item \textbf{Answer Attention.}
    We rank visual tokens by the direct attention assigned to them when predicting final-answer tokens.
    Unlike visual attribution, this signal does not trace attention through intermediate reasoning tokens.


    \item \textbf{TAM.}
    Token Activation Map~\citep{li2025tam} first projects visual hidden states onto the output classifier weight vector of each target text token and retains positive activations.
    It then estimates an interference map by aggregating the activation maps of preceding text tokens, weighted by their relevance to the target token, while excluding tokens identical to the target.
    This interference map is subtracted from the target-token map with a scale fitted by least squares.
    The resulting map is processed with a rank Gaussian filter and positive clipping to produce the final token-specific visual activation map.
    We use the resulting visual map to rank visual tokens.

    \item \textbf{Random.}
    We sample visual tokens uniformly without replacement,
    using the same 10\% selection budget.

    \item \textbf{Failure-Specific Contrastive VEP.}
    We rank visual tokens by
    $\mathbf{D}_Q^{-}
    = \boldsymbol{\mu}_Q^{-}-\boldsymbol{\mu}_Q^{+}$
    and select the highest-scoring positions.
    This emphasizes visual evidence attributed more strongly to failed rollouts than to successful rollouts.
\end{itemize}

\noindent\textbf{Pilot~3: Selecting response tokens.}
Pilot~3 compares Success-Specific and Failure-Specific Contrastive VEP and Success and Failure VEP selectors with VPPO, VGPO, Reasoning Attention, Token Entropy, and Random Selection.
All selectors operate on the same eligible response positions,
excluding final-answer tokens and special tokens.

\begin{itemize}
    \item \textbf{Success-Specific and Failure-Specific Contrastive VEP.}
    We use $\mathbf{D}_Q^{+}$ for correct rollouts and
    $\mathbf{D}_Q^{-}$ for incorrect rollouts to select
    visual states for replacement.
    Under the same response prefix, we measure the change
    in the log-probability of each observed response token:
    \[
        \Delta_{r,t}(\mathbf{D})
        =
        \log p_{r,t}
        -
        \log \widetilde{p}_{r,t}(\mathbf{D}),
    \]
    where $p_{r,t}$ and $\widetilde{p}_{r,t}(\mathbf{D})$
    are the probabilities before and after visual
    intervention, respectively.
    We select the response token with the largest
    positive score.

    \item \textbf{Success and Failure VEP.}
    We use $\boldsymbol{\mu}_Q^{+}$ for correct rollouts
    and $\boldsymbol{\mu}_Q^{-}$ for incorrect rollouts,
    replacing the contrastive prototypes while retaining
    the same intervention and response-token scoring
    procedure.

    \item \textbf{VPPO.}
    Following VPPO~\citep{huang2025vppo}, we rank candidate
    response tokens by the change in their predictive
    distributions induced by random image masking,
    with the response prefix held fixed.
    This measures sensitivity to image perturbation
    without outcome-specific visual selection.

    \item \textbf{VGPO.}
    Following VGPO~\citep{wang2026vgpo}, we rank candidate response tokens using hidden-state similarity to an aggregated visual representation.
    This measures visual alignment in representation space.

    \item \textbf{Reasoning Attention.}
    We rank candidate response tokens by their direct attention to visual tokens, without using outcome prototypes or visual interventions for scoring.

    \item \textbf{Token Entropy.}
    We rank candidate positions by the entropy of the
    next-token predictive distribution under the original
    image and response prefix, selecting the position
    with the highest entropy.
    This measures predictive uncertainty rather than
    visual dependency.

    \item \textbf{Random.}
    We select one response-token position uniformly
    from the same eligible candidate set.
\end{itemize}

Every response-token selector is evaluated with the same
retain-or-suppress continuation experiment.
Starting from the original prefix, the retain condition
forces the selected token for one decoding step, whereas
the suppress condition forbids that token at the same step.
For correct rollouts, we report
$\mathrm{Acc}(\mathrm{retain})-\mathrm{Acc}(\mathrm{suppress})$;
for incorrect rollouts, we report
$\mathrm{Acc}(\mathrm{suppress})-\mathrm{Acc}(\mathrm{retain})$.

\subsection{Success and Failure VEPs Distinguish Rollout Outcomes}
\label{app:pilot1_details}

For each question, Pilot~1 samples two sets of eight rollouts with independent
seeds. One set constructs the Success and Failure VEPs; the other is scored by
their relative cosine similarity. We then swap the two sets. We retain an
assignment if its construction set has at least two correct and two incorrect
rollouts and its evaluation set has both outcomes. Qwen2.5-VL contributes 750
questions and 9,456 evaluation responses. The same procedure applies to
Qwen3-VL and Qwen3.5. Sampling uses temperature \(1.0\), top-\(p=1.0\), and a
4096-token response cap.

Evaluation outcomes enter only the AUROC calculation. For Reasoning Attention,
we average total attention to visual tokens over all response positions.
The 95\% CIs use 10,000 question-level bootstrap samples.

\begin{table}[H]
    \centering
    \small
    \setlength{\tabcolsep}{3pt}
    \caption{Pilot~1 AUROC with 95\% confidence intervals.}
    \label{tab:pilot1_details}
    \begin{tabular}{@{}p{0.18\columnwidth}p{0.39\columnwidth}cp{0.24\columnwidth}@{}}
        \toprule
        Model & Signal & AUROC & 95\% CI \\
        \midrule
        \multirow{5}{*}{Qwen2.5-VL}
        & Success--Failure prototype contrast & \textbf{0.6845} & [0.6253, 0.7401] \\
        & Success VEP only & 0.6533 & [0.5968, 0.7086] \\
        & Answer Attention & 0.5209 & [0.4566, 0.5828] \\
        & Reasoning Attention & 0.5931 & [0.5356, 0.6481] \\
        & Shuffled support outcomes & 0.4957 & [0.4883, 0.5028] \\
        \addlinespace
        \multirow{5}{*}{Qwen3-VL}
        & Success--Failure prototype contrast & \textbf{0.6917} & [0.6371, 0.7437] \\
        & Success VEP only & 0.6316 & [0.5782, 0.6835] \\
        & Answer Attention & 0.5466 & [0.4875, 0.6040] \\
        & Reasoning Attention & 0.4854 & [0.4298, 0.5397] \\
        & Shuffled support outcomes & 0.4955 & [0.4871, 0.5033] \\
        \addlinespace
        \multirow{5}{*}{Qwen3.5}
        & Success--Failure prototype contrast & \textbf{0.7618} & [0.7157, 0.8044] \\
        & Success VEP only & 0.7023 & [0.6567, 0.7461] \\
        & Answer Attention & 0.5648 & [0.5135, 0.6127] \\
        & Reasoning Attention & 0.5333 & [0.4891, 0.5783] \\
        & Shuffled support outcomes & 0.4988 & [0.4911, 0.5065] \\
        \bottomrule
    \end{tabular}
\end{table}

The two-prototype score gives the highest AUROC on each model, while shuffled
outcomes remain near chance.

\subsection{Contrastive VEPs Identify Supporting Visual Tokens}
\label{app:pilot2_details}

Pilot~2 uses 950, 880, and 560 questions for Qwen2.5-VL, Qwen3-VL, and Qwen3.5,
respectively. Each selector ranks visual tokens, and we replace the top 10\%
of their hidden states with the mean visual state of the same image before the
first decoder layer. For Answer Attention, the ranking uses direct attention
from final-answer predictors to visual tokens. We then generate eight complete
responses per question at temperature \(1.0\) and top-\(p=1.0\). The response cap
is 4096 tokens for Qwen2.5-VL and 2048 for Qwen3-VL and Qwen3.5.

Accuracy drop is measured against clean generation. We average the eight
response accuracies for each question and use the same aggregation for every
selector. The 95\% CIs use 20,000 question-level, benchmark-stratified bootstrap
samples.

\begin{table}[H]
    \centering
    \footnotesize
    \setlength{\tabcolsep}{2.5pt}
    \caption{Pilot~2 accuracy drop in percentage points. \(\dagger\) marks a
    paired 95\% CI above zero.}
    \label{tab:pilot2_details}
    \begin{tabular}{@{}p{0.38\columnwidth}ccc@{}}
        \toprule
        Selector & Qwen2.5-VL & Qwen3-VL & Qwen3.5 \\
        \midrule
        \(\mathbf D_Q^+\) & \textbf{28.45} & \textbf{20.61} & \textbf{20.15} \\
        \quad 95\% CI & [22.34, 34.54] & [14.02, 27.68] & [10.57, 29.08] \\
        \(\boldsymbol\mu_Q^+\) & 26.71 & 17.71 & 11.93 \\
        Saliency-R1 & 18.81 & 9.13 & 8.93 \\
        Answer Attention & 17.06 & -0.42 & 11.96 \\
        TAM & 5.60 & -0.91 & 1.49 \\
        Random & 3.09 & 0.57 & 2.08 \\
        \quad 95\% CI & [-1.21, 7.34] & [-4.17, 5.02] & [-2.86, 6.99] \\
        \(\mathbf D_Q^-\) & 0.77 & -3.20 & -4.17 \\
        \midrule
        Extra drop vs. Random & 25.36\(^{\dagger}\) & 20.04\(^{\dagger}\) & 18.07\(^{\dagger}\) \\
        \quad 95\% CI & [18.72, 32.07] & [12.83, 27.21] & [8.27, 27.44] \\
        \bottomrule
    \end{tabular}
\end{table}

\(\mathbf D_Q^+\) produces the largest drop on all three models. Its extra
drop over Random ranges from 18.07 to 25.36 points.

\subsection{Contrastive VEPs Locate Supporting and Harmful Response Tokens}
\label{app:pilot3_details}

Pilot~3 uses groups of eight with at least three correct and three incorrect
rollouts, selecting one of each for intervention. The VEPs are computed from
the full group. After replacing the \(\mathbf D_Q^+\)- or
\(\mathbf D_Q^-\)-selected visual states as in Pilot~2, we rank non-special
response tokens with nonempty decoded text by \(\Delta_{r,t}\) in
Eq.~\eqref{eq:token_logp_change}. We test all eligible positions and then
exclude final-answer tokens.

We retain the token with the largest positive score at its original prefix or
suppress it for one step. Both conditions sample eight continuations with the
same question-level seed, temperature \(1.0\), top-\(p=1.0\), and a 4096-token
cap. The effect is \(\operatorname{Acc}(\mathrm{retain})-
\operatorname{Acc}(\mathrm{suppress})\) for correct rollouts and the reverse for
incorrect rollouts. The 95\% CIs use 20,000 question-level,
benchmark-stratified bootstrap samples.

\begin{table}[H]
    \centering
    \footnotesize
    \setlength{\tabcolsep}{2pt}
    \caption{Pilot~3 intervention effects in accuracy points, with 95\% CIs.}
    \label{tab:pilot3_details}
    \begin{tabular}{@{}lrlrlrl@{}}
        \toprule
        & & & \multicolumn{2}{c}{Correct rollouts}
        & \multicolumn{2}{c}{Incorrect rollouts} \\
        \cmidrule(lr){4-5}\cmidrule(lr){6-7}
        Model / checkpoint & Questions & Response tokens
        & Effect & 95\% CI & Effect & 95\% CI \\
        \midrule
        \multirow{2}{*}{Qwen2.5-VL / PAPO}
        & \multirow{2}{*}{3,170} & All
        & \(+16.17\) & [12.07, 20.43] & \(+12.85\) & [8.95, 16.88] \\
        & & Excl. final answer
        & \(+8.71\) & [5.60, 11.95] & \(+4.10\) & [1.46, 6.86] \\
        \addlinespace
        \multirow{2}{*}{Qwen3-VL / VPPO-8B}
        & \multirow{2}{*}{1,350} & All
        & \(+5.65\) & [1.48, 10.00] & \(+3.61\) & [0.47, 7.78] \\
        & & Excl. final answer
        & \(+5.65\) & [1.39, 10.00] & \(+3.61\) & [0.59, 7.87] \\
        \addlinespace
        \multirow{2}{*}{Qwen3.5-9B / official}
        & \multirow{2}{*}{1,730} & All
        & \(+11.56\) & [7.37, 15.97] & \(+11.13\) & [6.94, 15.39] \\
        & & Excl. final answer
        & \(+10.04\) & [6.07, 14.16] & \(+11.27\) & [7.08, 15.53] \\
        \bottomrule
    \end{tabular}
\end{table}

After excluding final-answer tokens, all six effects have positive 95\% CIs.
\FloatBarrier

\onecolumn
\section{Theoretical Analysis}
\label{app:theory}

\subsection{Contrastive Prototype Recovery}
\label{app:prototype_recovery}

\paragraph{Attribution model.}
Fix an image--question pair \(Q\) and the sampling policy, and omit \(Q\)
from the notation. For outcome \(y\in\{+,-\}\), assume the normalized
attribution of rollout \(r\) admits the decomposition
\begin{equation}
    \mathbf v_r=\mathbf b+\mathbf u^y+\boldsymbol\epsilon_r,
    \qquad
    \|\mathbf b\|_1=1-\alpha,
    \qquad
    \|\mathbf u^y\|_1=\alpha,
    \label{eq:theory_attribution_model}
\end{equation}
where \(0<\alpha<1\), \(\mathbf b,\mathbf u^+,\mathbf u^-\) are
nonnegative, and the supports of \(\mathbf u^+\) and \(\mathbf u^-\) are
disjoint. The common component \(\mathbf b\) can overlap either support.
The noise has conditional mean zero and zero coordinate sum, so
\(\mathbf v_r\) remains in the probability simplex. We apply this model after
the normalization in Eq.~\eqref{eq:rollout_visual_attribution}. The components
\(\mathbf b,\mathbf u^+,\mathbf u^-\) belong to the assumed data-generating
model and are not defined from \(\mathbf D^\pm\).

Let \(n_y>0\) be the number of rollouts with outcome \(y\), and assume
independent samples within each outcome. Each noise coordinate is
sub-Gaussian with parameter \(\sigma_y^2\), meaning
\(\mathbb E[\exp(t\epsilon_{r,j})\mid y]
\le\exp(t^2\sigma_y^2/2)\) for all real \(t\).
For \(d=|\mathcal I_Q|\) visual positions and \(\delta\in(0,1)\), define
\begin{equation}
    e_y=\sigma_y\sqrt{\frac{2\log(4d/\delta)}{n_y}},
    \qquad e=e_++e_-.
    \label{eq:theory_prototype_error}
\end{equation}
The sample means satisfy
\(\boldsymbol\mu^y=\mathbf b+\mathbf u^y+
\overline{\boldsymbol\epsilon}^{\,y}\).
Concentration of each sample mean and a union bound over the two groups and
\(d\) positions give
\begin{equation}
    \Pr\!\left(
    \|\overline{\boldsymbol\epsilon}^{\,+}\|_\infty\le e_+,
    \ \|\overline{\boldsymbol\epsilon}^{\,-}\|_\infty\le e_-
    \right)\ge 1-\delta.
    \label{eq:theory_error_event}
\end{equation}
When an independent rollout group is partitioned by outcome, this probability
statement is conditional on the observed counts \(n_+\) and \(n_-\).

\begin{proposition}[Contrastive prototype recovery]
\label{prop:prototype_recovery}
Under the attribution model above, let
\(S^y=\operatorname{supp}(\mathbf u^y)\), with \(e\) defined in
Eq.~\eqref{eq:theory_prototype_error}. With probability at least
\(1-\delta\),
\begin{equation}
    \min_{j\in S^y}u_j^y>2e
    \quad\Longrightarrow\quad
    \operatorname{Top}_{|S^y|}(\mathbf D^y)=S^y.
    \label{eq:prototype_support_recovery}
\end{equation}
\end{proposition}

\paragraph{Proof of Proposition~\ref{prop:prototype_recovery}.}
We prove the positive case; the negative case follows by exchanging outcomes.
Let \(S^+=\operatorname{supp}(\mathbf u^+)\), \(m=|S^+|\), and
\(u_{\min}=\min_{j\in S^+}u_j^+>2e\).
On the event in Eq.~\eqref{eq:theory_error_event}, subtracting the prototypes
eliminates the common component:
\begin{equation}
    \boldsymbol\mu^+-\boldsymbol\mu^-
    =\mathbf u^+-\mathbf u^-
    +\overline{\boldsymbol\epsilon}^{\,+}
    -\overline{\boldsymbol\epsilon}^{\,-}.
    \label{eq:theory_common_cancellation}
\end{equation}
For \(j\in S^+\), disjoint support gives \(u_j^-=0\), so
\(D_j^+\ge u_{\min}-e>e\). For \(j\notin S^+\), the population
difference is nonpositive, so \(D_j^+\le e\). Every target position
therefore ranks above every other position, establishing exact Top-\(m\)
recovery with probability at least \(1-\delta\). The comparison does not
depend on the shape or peak value of \(\mathbf b\).

The raw Success VEP can instead rank only common positions. To see this,
suppose \(d\ge3m\), take three disjoint sets \(B,S^+,S^-\), each of size
\(m\), and choose
\begin{equation}
    b_j=\frac{1-\alpha}{m}\mathbf1[j\in B],
    \qquad
    u_j^\pm=\frac{\alpha}{m}\mathbf1[j\in S^\pm],
    \qquad 0<\alpha<\frac12.
    \label{eq:theory_raw_counterexample}
\end{equation}
Without noise, every common position has raw attribution
\((1-\alpha)/m>\alpha/m\). Thus, the Top-\(m\) positions of
\(\boldsymbol\mu^+\) are exactly \(B\), whereas those of
\(\mathbf D^+\) are exactly \(S^+\). The two rankings remain different on the
finite-sample event whenever
\begin{equation}
    \frac{\alpha}{m}>2(e_++e_-),
    \qquad
    \frac{1-2\alpha}{m}>2e_+.
    \label{eq:theory_raw_failure_margin}
\end{equation}
For any fixed \(\alpha\in(0,1/2)\), both inequalities hold for sufficiently
small nonzero noise scales. The counterexample therefore does not rely on a
noiseless limit. \hfill\(\square\)

\paragraph{Interpretation.}
The proposition guarantees recovery of the outcome component in
Eq.~\eqref{eq:theory_attribution_model}. Common attention may still be useful for reasoning, but its
magnitude need not identify the contrastive target. Subtraction also carries
estimation error from both outcome groups, so recovery depends on both the
remaining signal and within-outcome variation. Here \(m\) counts visual
positions.

\subsection{Local Policy Improvement}
\label{app:local_improvement}

\paragraph{Update model.}
Let \(J(\theta)\) be expected return and fix the current parameter
\(\theta_0\). Decompose a sampled policy-gradient update direction into
selected and remaining response-token contributions:
\begin{equation}
    H=H_{\mathcal K}+H_{\mathcal N},
    \qquad
    H_w=aH_{\mathcal K}+bH_{\mathcal N}.
    \label{eq:theory_gradient_split}
\end{equation}
The contributions include the signed advantages and the original averaging
factors. We hold the response and selection sizes fixed, so
\(q=|\mathcal K_i|/|\mathcal U_i|\) is constant. The weights
\(a=1+\lambda_{\mathrm{tok}}(1-q)\) and
\(b=1-\lambda_{\mathrm{tok}}q\) are exactly those in
Eq.~\eqref{eq:token_vep_guidance}. We assume
\(0<q<1\) and \(0<\lambda_{\mathrm{tok}}<1/q\), ensuring \(0<b<a\).
Both directions use the same trajectory-level guidance. We compare the local
policy-gradient component before clipping and omit optimizer preconditioning
and the full KL-to-reference dynamics.

\paragraph{Sparse ascent and gradient separation.}
Assume the selected contribution has a mean ascent direction \(\mathbf h\),
while the remaining contribution has zero mean:
\begin{equation}
    \mathbb E H_{\mathcal K}=\mathbf h,
    \qquad \mathbb E H_{\mathcal N}=0,
    \qquad \nabla J(\theta_0)^\top\mathbf h>0.
    \label{eq:theory_sparse_ascent}
\end{equation}
Let \(F\succeq0\) be the Fisher matrix at \(\theta_0\), under a fixed
old-policy state distribution. Assume
\begin{equation}
    \mathbb E[H_{\mathcal K}^\top F H_{\mathcal N}]=0,
    \qquad
    V_{\mathcal K}=\mathbb E[H_{\mathcal K}^\top F H_{\mathcal K}]>0,
    \qquad
    V_{\mathcal N}=\mathbb E[H_{\mathcal N}^\top F H_{\mathcal N}]>0,
    \label{eq:theory_fisher_moments}
\end{equation}
with finite moments. VPPO uses a related gradient decomposition to motivate
visual-token filtering~\citep{huang2025vppo}. Here we use the split only to
compare selected and remaining response-token contributions under the same
quadratic KL budget.

\begin{proposition}[Local policy improvement]
\label{prop:local_improvement}
Under the assumptions above, the leading-order return gains at the same
expected quadratic KL budget satisfy
\begin{equation}
    \frac{\Delta J_{\mathrm{VEP}}^{(1)}}
         {\Delta J_{\mathrm{uniform}}^{(1)}}
    =
    \sqrt{\frac{V_{\mathcal K}+V_{\mathcal N}}
    {V_{\mathcal K}+(b/a)^2V_{\mathcal N}}}
    >1.
    \label{eq:local_improvement_ratio}
\end{equation}
\end{proposition}

\paragraph{Proof of Proposition~\ref{prop:local_improvement}.}
For a parameter displacement \(u\), averaging over the fixed state
distribution \(d_{\mathrm{old}}\) gives
\begin{equation}
    \mathbb E_{s\sim d_{\mathrm{old}}}
    D_{\mathrm{KL}}\!\left(
    \pi_{\theta_0}(\cdot\mid s)\,\|\,
    \pi_{\theta_0+u}(\cdot\mid s)\right)
    =\tfrac12u^\top Fu+o(\|u\|^2).
    \label{eq:theory_kl_expansion}
\end{equation}
Given a budget \(\kappa>0\), choose comparison step sizes
\begin{equation}
    \eta_0=\sqrt{\frac{2\kappa}{V_{\mathcal K}+V_{\mathcal N}}},
    \qquad
    \eta_w=\sqrt{\frac{2\kappa}
    {a^2V_{\mathcal K}+b^2V_{\mathcal N}}}.
    \label{eq:theory_matched_steps}
\end{equation}
Equation~\eqref{eq:theory_fisher_moments} implies
\begin{equation}
    \tfrac12\mathbb E[(\eta_0H)^\top F(\eta_0H)]
    =\tfrac12\mathbb E[(\eta_wH_w)^\top F(\eta_wH_w)]
    =\kappa.
    \label{eq:theory_equal_budget}
\end{equation}
The two directions therefore have the same expected quadratic KL cost. We use
this normalization only for the theoretical comparison; VEP-RL does not add a
learning-rate controller.

Writing \(c=\nabla J(\theta_0)^\top\mathbf h>0\),
Eq.~\eqref{eq:theory_sparse_ascent} gives
\begin{equation}
    \Delta J_{\mathrm{uniform}}^{(1)}=\eta_0c,
    \qquad
    \Delta J_{\mathrm{VEP}}^{(1)}=a\eta_wc.
    \label{eq:theory_first_order_gains}
\end{equation}
Substituting Eq.~\eqref{eq:theory_matched_steps} yields
Eq.~\eqref{eq:local_improvement_ratio}. The inequality is strict because
\(b/a<1\) and \(V_{\mathcal N}>0\). \hfill\(\square\)

\paragraph{Finite local updates.}
Suppose the gradient of \(J\) is \(L\)-Lipschitz along all realized update
segments, and both directions have finite Euclidean second moments. Define
\(M_0=\mathbb E\|H\|^2/(V_{\mathcal K}+V_{\mathcal N})\) and
\(M_w=\mathbb E\|H_w\|^2/(a^2V_{\mathcal K}+b^2V_{\mathcal N})\).
Taylor bounds for the two updates give
\begin{equation}
    \mathbb E[J(\theta_0+\eta_wH_w)-J(\theta_0+\eta_0H)]
    \ge C\sqrt{\kappa}-L\kappa(M_0+M_w),
    \label{eq:theory_finite_gain}
\end{equation}
where
\begin{equation}
    C=\sqrt2\,c\left[
    \frac{a}{\sqrt{a^2V_{\mathcal K}+b^2V_{\mathcal N}}}
    -\frac{1}{\sqrt{V_{\mathcal K}+V_{\mathcal N}}}
    \right]>0.
    \label{eq:theory_finite_constant}
\end{equation}
The bound is positive for sufficiently small \(\kappa\). The comparison
matches quadratic KL costs, not exact finite-step KL divergences.

\paragraph{Scope.}
The sparse-ascent condition in Eq.~\eqref{eq:theory_sparse_ascent} is not
implied by Proposition~\ref{prop:prototype_recovery}. Spatial recovery and
response-token selection are different steps. Pilot~3 tests response-token
selection through answer accuracy. If the selector misses a contribution with
nonzero mean, or if cross terms provide important noise cancellation, the
improvement need not hold. The proposition gives a local comparison under the
stated assumptions; it does not establish convergence of the full
shared-parameter training process.

\begingroup
\makeatletter
\setlength{\@fptop}{0pt}
\setlength{\@fpsep}{12pt}
\setlength{\@fpbot}{0pt plus 1fil}
\makeatother
\onecolumn
\section{Additional Analysis}
\label{app:additional_analysis}
\subsection{Error Analysis}
\label{app:error_analysis}

\paragraph{Error analysis.}
\begin{wrapfigure}{R}{0.50\textwidth}
    \centering
    \includegraphics[width=\linewidth]{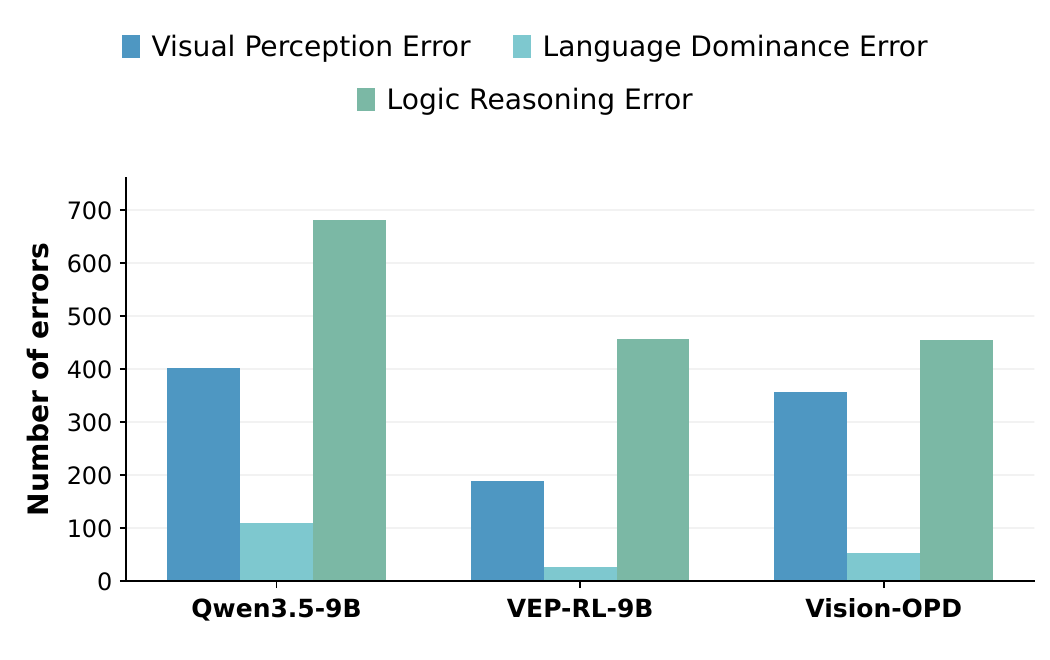}
    \caption{Error breakdown by category.}
    \label{fig:failure_modes}
\end{wrapfigure}

To understand when VEP-RL fails, we analyze 3,000 sampled responses from Qwen3.5-9B, Vision-OPD, and VEP-RL-9B on evaluation benchmarks and classify each incorrect response. We observe three failure patterns as shown in Figure~\ref{fig:failure_modes}: (i) Visual Perception Errors, where the model misperceives visual details;  
(ii) Language Dominance Error, where correctly perceived visual evidence is subjected to language priors; 
and (iii) Logic Reasoning Errors, where the model correctly extracts the visual information but reasons incorrectly. 
We first used GPT-5.6 Sol to assign a preliminary category to each incorrect response based on its image, question, ground-truth answer, and model rollout, and then manually reviewed the assignments to determine the final labels. 
The error breakdown indicates that VEP-RL first improves visual perception compared with Vision-OPD and the base model. Second, VEP-RL reduces language-dominance errors, potentially because VEP-RL explicitly reinforces visually supported tokens. 


\subsection{Selected-token Frequencies}
\label{app:token_frequencies}


We showcase what VEP selects in Pilot ~\ref{sec:pilot_token_intervention}. Figure~\ref{fig:token_wordclouds} combines the tokens selected by
\(\mathbf D_Q^+\) and \(\mathbf D_Q^-\) across the three model-specific sets. We remove
leading and trailing whitespace. We also omit whitespace-only strings,
Unicode replacement and control characters, tag or \LaTeX{} command fragments
identified in context, and punctuation-only strings except \%.

\begin{figure}[H]
    \centering
    \includegraphics[width=\linewidth]{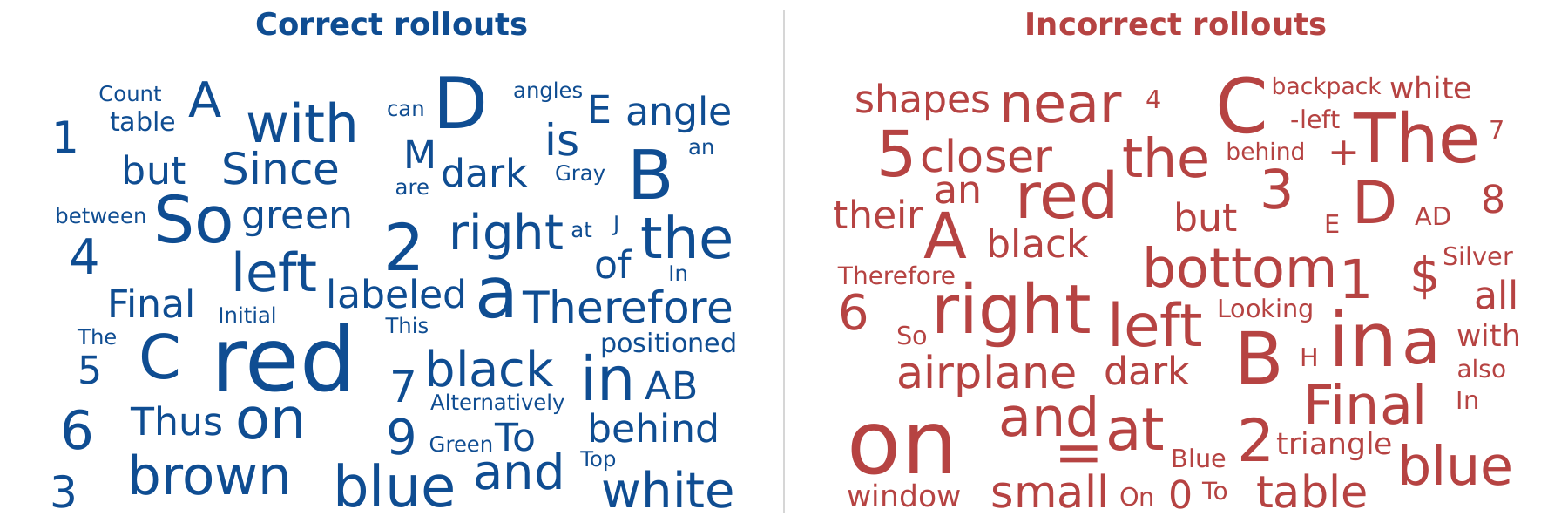}
    \caption{Word clouds of the Top-1 response tokens selected by contrastive
    VEPs. Text size indicates frequency within each panel, not intervention
    effect, and is not comparable across panels.}
    \label{fig:token_wordclouds}
\end{figure}

\FloatBarrier

\endgroup
\onecolumn
\subsection{Case Study}
\label{app:case_study}

{\centering
\includegraphics[
    width=0.98\linewidth,
    height=0.84\textheight,
    keepaspectratio
]{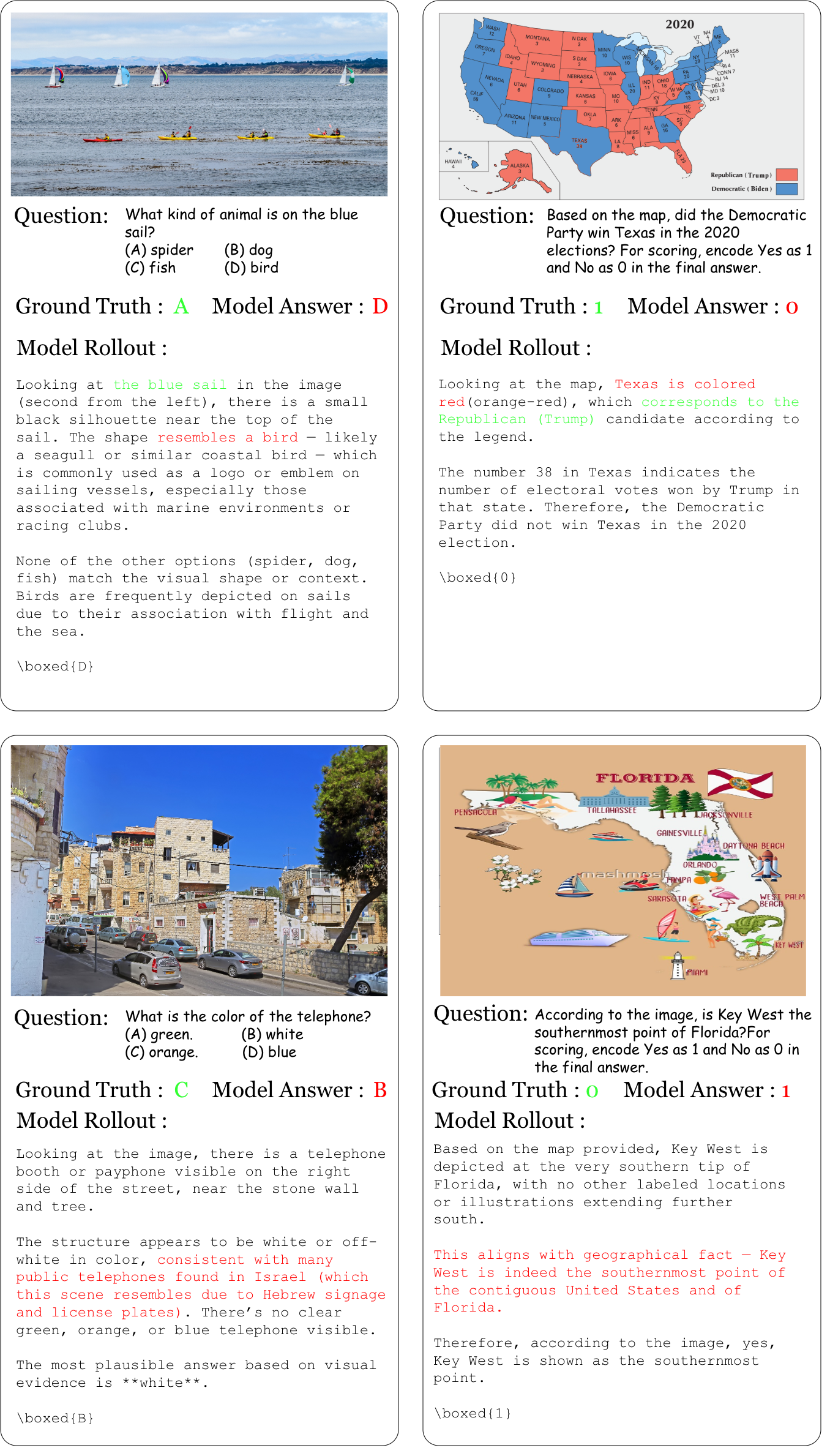}
\par}

\captionof{figure}{
\textbf{Visual Perception Errors and Language Dominance Errors.}
\textbf{Top left:} The model mistakes the small spider emblem on the blue sail for a bird.
\textbf{Top right:} It reads the blue fill of Texas as red and consequently identifies the wrong winning party.
\textbf{Bottom left:} An assumption about the typical appearance of public telephones overrides the target object's visual evidence.
\textbf{Bottom right:} The model relies on its knowledge of Key West rather than the spatial arrangement depicted in the illustration.
}
\label{fig:case_study_1}

{\centering
\includegraphics[
    width=0.98\linewidth,
    height=0.84\textheight,
    keepaspectratio
]{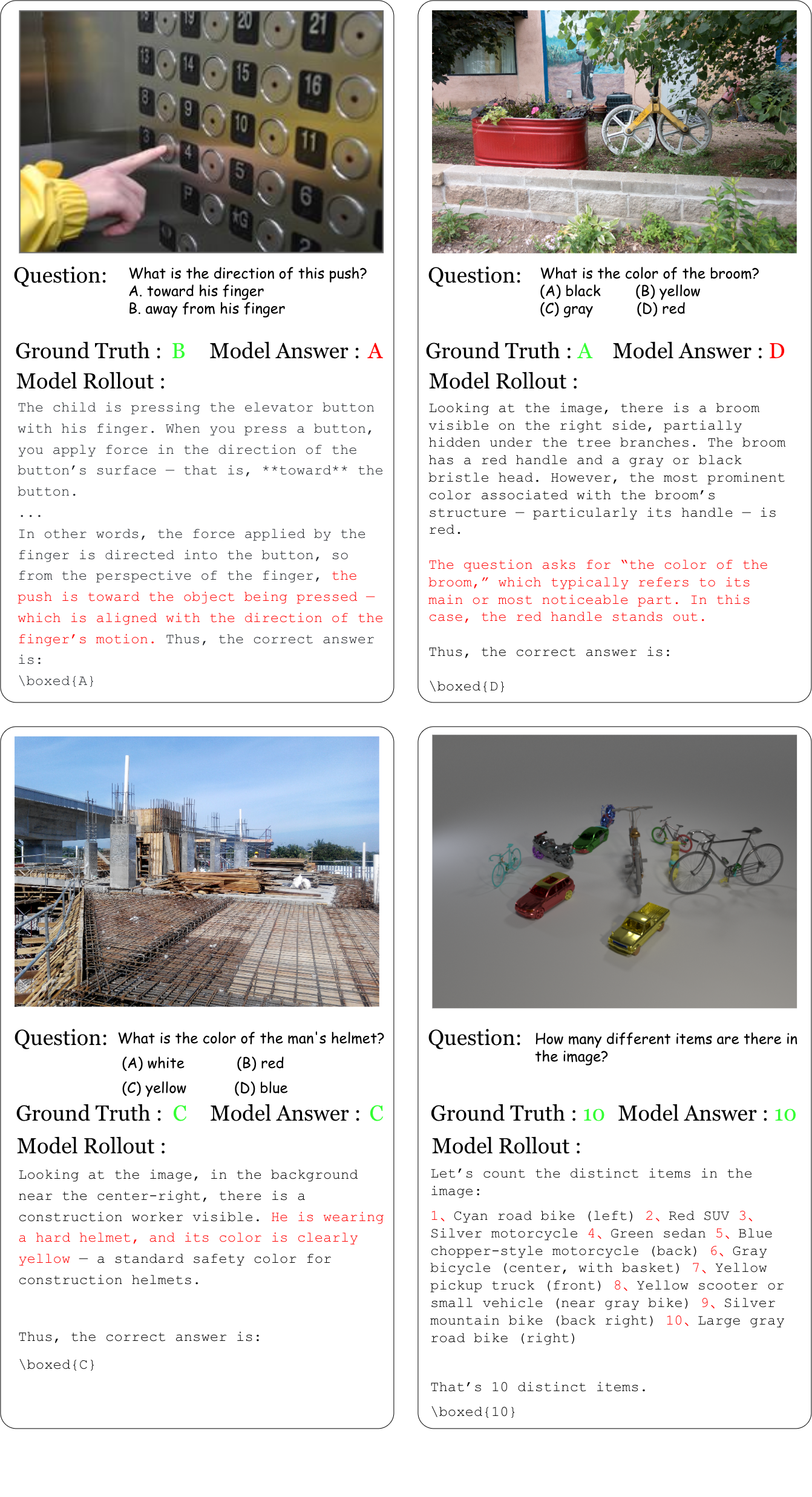}
\par}

\captionof{figure}{
\textbf{Logic Reasoning Errors and Successful Cases.}
\textbf{Top left:} The model recognizes that the finger pushes into the elevator button but reverses the direction relative to the finger.
\textbf{Top right:} It identifies the broom's red handle and dark bristles, yet incorrectly selects the handle as the broom's defining color.
\textbf{Bottom left:} The model locates the construction worker and correctly identifies the yellow helmet.
\textbf{Bottom right:} It enumerates the objects in the scene and correctly counts ten distinct items.
}
\label{fig:case_study_2}

\begin{figure}[tbp]
    \centering
    \includegraphics[width=0.95\linewidth]{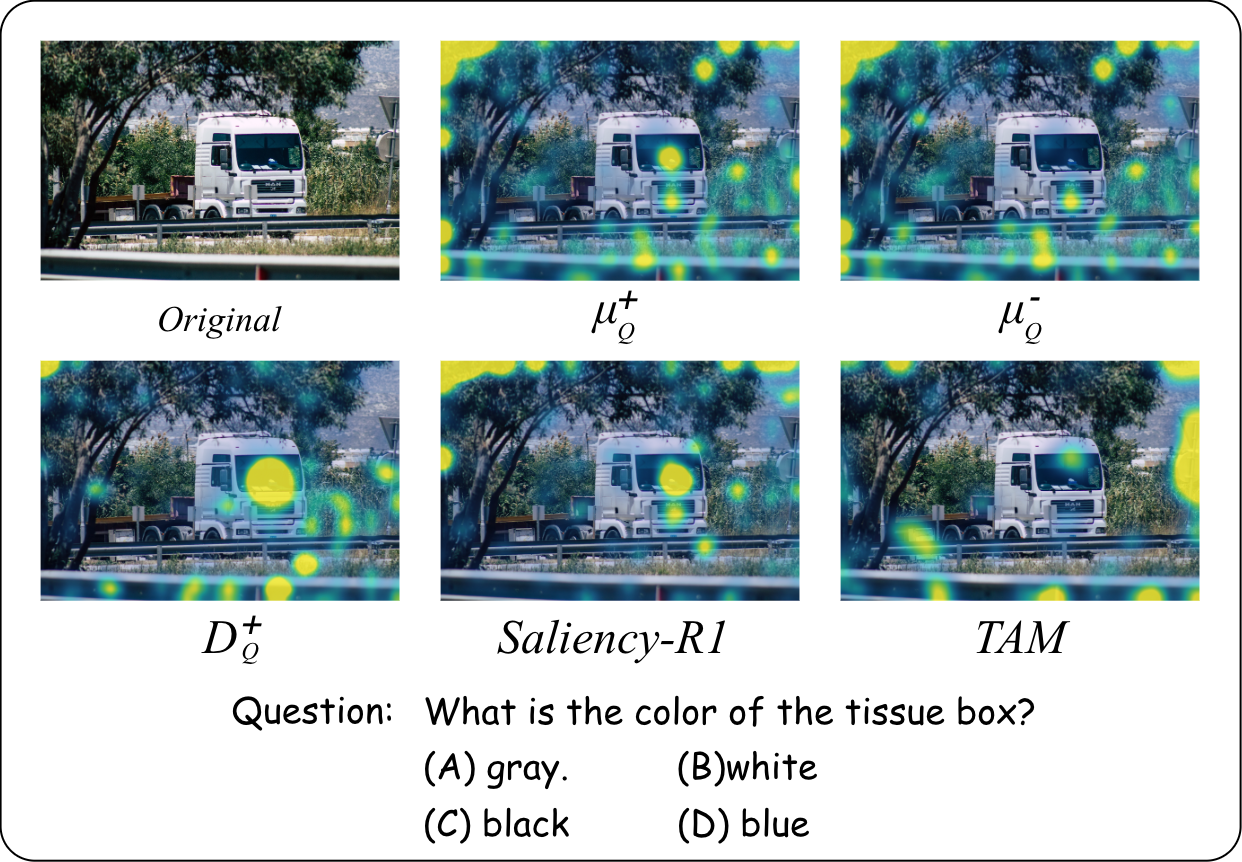}
    \caption{The tissue box is a small object inside the truck windshield. Both our contrastive map $D_Q^{+}$ and Saliency-R1 highlight the windshield region, while TAM's most prominent responses appear elsewhere in the scene.}
    \label{fig:heatmap_tissue_box}
\end{figure}

\begin{figure}[tbp]
    \centering
    \includegraphics[width=0.95\linewidth]{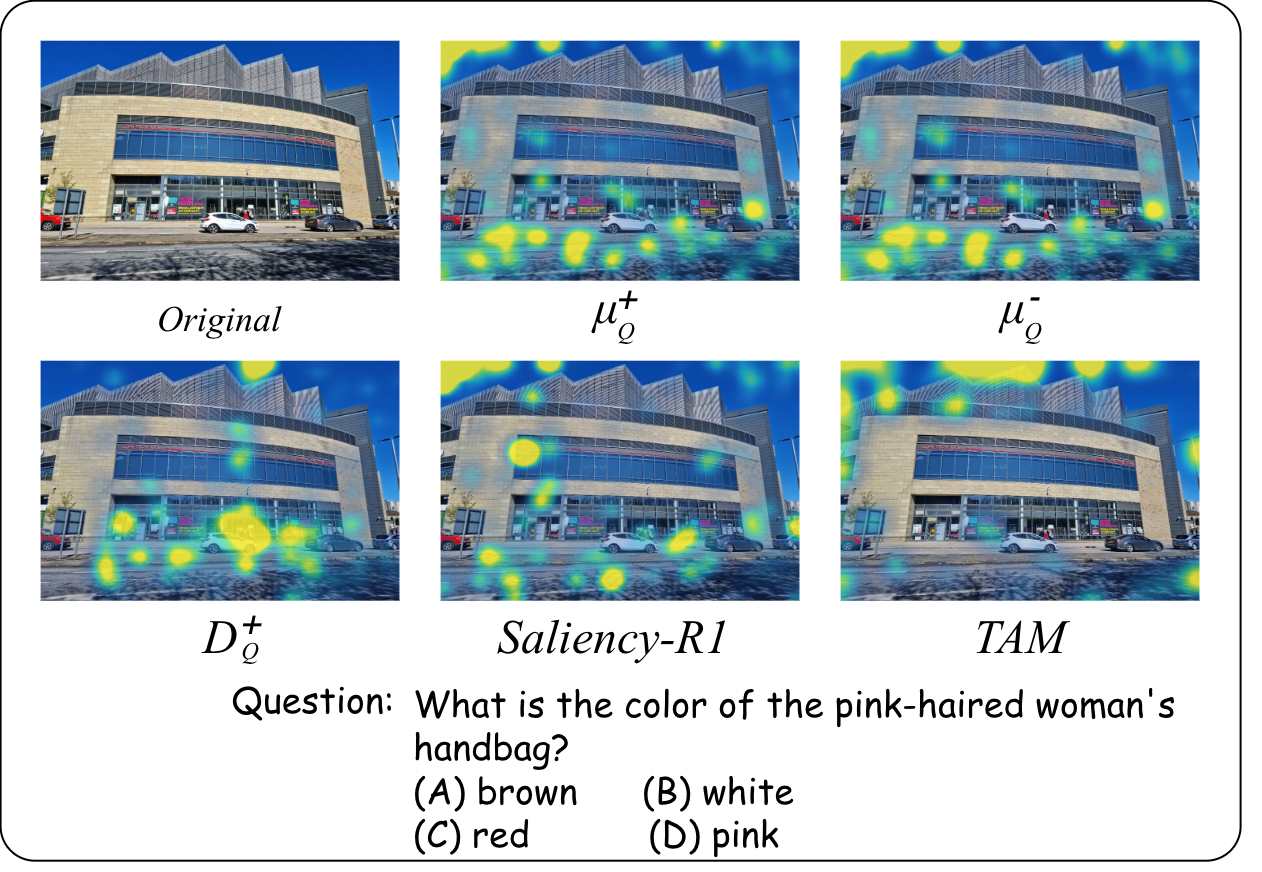}
    \caption{The pink-haired woman is located near the white car in front of the building. Only our contrastive map $D_Q^{+}$ clearly highlights her location; Saliency-R1 and TAM focus mainly on other regions.}
    \label{fig:heatmap_handbag}
\end{figure}





\end{document}